\documentclass[final,12pt]{clear2023} 

\usepackage{siunitx} 

\usepackage{xcolor}         
\definecolor{citecolor}{HTML}{0071bc}
\usepackage{mathrsfs}
\usepackage{booktabs}       
\usepackage{amsfonts}       
\usepackage{nicefrac}       
\usepackage{microtype}      
\usepackage{soul}
\usepackage{xfrac}
\usepackage{tikz,pgf}
\usepackage{mathtools}
\usepackage{thmtools,thm-restate}
\usepackage{bbm}
\usepackage{dsfont}

\def\*#1{\mathbf{#1}}

\usepackage{wrapfig}

\usetikzlibrary{positioning,shapes,shapes.misc,arrows,calc,arrows.meta,fit, backgrounds,quotes,intersections,patterns,decorations.pathmorphing,decorations.pathreplacing,decorations.markings}
\tikzset{
  >={Latex[width=1.5mm,length=1.7mm]},
  RR/.style={draw,circle,inner sep=0mm, minimum size=4.mm,font=\sffamily\footnotesize},
  intervened/.style={RR,draw=red,red,thin},
  AUX/.style={draw=none,inner sep=0mm,font=\sffamily\scriptsize},
  unobserved/.style={RR,fill=black!40,text=white,draw=black!40},
  every picture/.style=semithick,
  snakey/.style={decorate, decoration={snake,segment length=2mm, amplitude=.25mm,pre length=.5mm, post length=.5mm}},
  given/.style={fill=gray!50},
  ggiven/.style={RR,draw=bettergreen,thick,fill=bettergreen!50},
}

\newtheorem{re-lemma}[theorem]{Lemma}
\newtheorem{re-proposition}[theorem]{Proposition}
\newtheorem{re-remark}[theorem]{Remark}
\newtheorem{re-corollary}[theorem]{Corollary}
\newtheorem{re-definition}[theorem]{Definition}
\newtheorem{re-conjecture}[theorem]{Conjecture}
\newtheorem{re-axiom}[theorem]{Axiom}

\usepackage{cleveref}
\Crefname{figure}{Fig.}{Figs.}
\Crefname{section}{Sec.}{Sec.}
\Crefname{equation}{Eq.}{Eqs.}
\Crefname{proposition}{Prop.}{Props.}
\Crefname{prop}{Prop.}{Props.}
\Crefname{theorem}{Thm.}{Thms}
\Crefname{lemma}{Lemma}{Lemmas}
\Crefname{definition}{Def.}{Defs}
\Crefname{algorithm}{Alg.}{Algs}
\Crefname{remark}{Remark}{Remarks}
\Crefname{example}{Example}{Examples}
\Crefname{corollary}{Corollary}{Corollary}
\Crefname{re-corollary}{Corollary}{Corollary}

\newcommand{\ours}{NCD}
\newcommand{\oursFull}{Neural Contextual Decomposition}

\newcommand{\oursfull}{neural contextual decomposition}

\newcommand{\newvariable}{partition indicator variable}

\newcommand{\nv}{PIV}
\newcommand{\comment}[1]{}
\newcommand{\cmnt}[1]{}

\newcommand{\newcsi}{context-set specific independence}

\newcommand{\NewCsi}{Context-Set Specific Independence}
\newcommand{\nc}{CSSI}

\newcommand{\ctxdec}{contextual decomposition}
\newcommand{\Ctxdec}{Contextual decomposition}
\newcommand{\CtxDec}{Contextual Decomposition}
\newcommand{\cd}{CD}

\usepackage{amsmath,amsfonts,bm}

\def\eqref#1{equation~\ref{#1}}

\def\1{\bm{1}}

\def\rvx{{\mathbf{x}}}

\def\rmX{{\mathbf{X}}}
\def\rmY{{\mathbf{Y}}}

\DeclareMathAlphabet{\mathsfit}{\encodingdefault}{\sfdefault}{m}{sl}
\SetMathAlphabet{\mathsfit}{bold}{\encodingdefault}{\sfdefault}{bx}{n}

\def\gD{{\mathcal{D}}}
\def\gE{{\mathcal{E}}}
\def\gF{{\mathcal{F}}}

\def\gI{{\mathcal{I}}}

\def\gX{{\mathcal{X}}}

\def\gZ{{\mathcal{Z}}}

\newcommand\Perp{\protect\mathpalette{\protect\independenT}{\perp}} \def\independenT#1#2{\mathrel{\rlap{$#1#2$}\mkern3mu{#1#2}}}

\title[Neural Contextual Decomposition]{On Discovery of Local Independence over Continuous Variables\\ via Neural Contextual Decomposition}

\usepackage{times}

\clearauthor{
 \Name{Inwoo Hwang} \Email{bluemoon1@snu.ac.kr}\\
 \addr AI Institute, Seoul National University
 \AND
 \Name{Yunhyeok Kwak} \Email{yunh$\_$kwak@snu.ac.kr}\\
 \addr AI Institute, Seoul National University
 \AND
 \Name{Yeon-Ji Song} \Email{yjsong@snu.ac.kr}\\
 \addr AI Institute, Seoul National University
 \AND
 \Name{Byoung-Tak Zhang}\thanks{Corresponding authors.} \Email{btzhang@snu.ac.kr}\\
 \addr AI Institute, Seoul National University
 \AND
 \Name{Sanghack Lee}\footnotemark[1] \Email{sanghack@snu.ac.kr}\\
 \addr Graduate School of Data Science, Seoul National University
}

\begin{document}

\maketitle

\begin{abstract}
Conditional independence provides a way to understand causal relationships among the variables of interest. An underlying system may exhibit more fine-grained causal relationships especially between a variable and its parents, which will be called the local independence relationships. One of the most widely studied local relationships is Context-Specific Independence (CSI), which holds in a specific assignment of conditioned variables. 
However, its applicability is often limited since it does not allow continuous variables: data conditioned to the specific value of a continuous variable contains few instances, if not none, making it infeasible to test independence.
In this work, we define and characterize the local independence relationship that holds in a specific set of joint assignments of parental variables, which we call \newcsi{} (\nc). 
We then provide a canonical representation of \nc{} and prove its fundamental properties.
Based on our theoretical findings, we cast the problem of discovering multiple \nc{} relationships in a system as finding a partition of the joint outcome space.
Finally, we propose a novel method, coined \oursfull{} (\ours), which learns such partition by imposing each set to induce \nc{} via modeling a conditional distribution.
We empirically demonstrate that the proposed method successfully discovers the ground truth local independence relationships in both synthetic dataset and complex system reflecting the real-world physical dynamics. 
\end{abstract}


\section{Introduction}
\label{sec:introduction}
Discovering the causal relationships in a system is an important and challenging problem in many areas of scientific research such as social science \citep{sobel1995causal}, biology \citep{shipley2016cause}, and economics  \citep{angrist1996identification,angrist2008mostly,imbens2015causal,banerjee2016influence,imbens2020potential}.
There have been many causal discovery algorithms finding causal relationships given observational data \citep{spirtes2000constructing, chickering2002optimal, hoyer2008nonlinear, shimizu2006linear, zhang2011kernel, zheng2018dags, zhu2019causal}. These methods often exploit conditional independence either explicitly (constraint-based methods) or implicitly (score-based or gradient-based).

A system often exhibits more fine-grained relationships between a variable and its parents, i.e., a local independence relationship. 
For example, when pushing an object on the ground, it will move \textit{only} when a force exceeds the friction which is determined by the mass and the ground.
Context-specific independence (CSI) \citep{poole2003exploiting,boutilier2013contextspecific} is the independence that holds in a \textit{certain} conditioning value (i.e., \textit{context}) as opposed to \textit{any} value of the conditioned variables. 
It has been shown that such independence leads to more efficient probabilistic inference by exploiting the underlying local structure \citep{poole2013contextspecific,poole2003exploiting,gogate2010exploiting,van2011lifted,dal2018parallel}.
Further, it allows the identification of causal effects, which would not be possible without CSI relationships \citep{tikka2020identifying, robins2020interventionist}. 

Despite the fact that many real-world scenarios involve continuous variables, most prior works on local independence relationships assumed that variables are discrete. This is partly due to the notion of CSI that is inherently suited for discrete variables as conditioning on the specific value of a continuous variable is infeasible, where the resulting subset of data would be practically empty. Hence, it is non-trivial to explore how local independence for continuous variables can be empirically discovered and utilized in probabilistic or causal inferential tasks. 

\begin{figure}
\subfigure[]{\label{fig:example-a}\centering
\begin{tikzpicture}[x=7mm,y=15mm]
    \node[RR] (X1) {$X_1$};
    \node[RR] (X2) at (2,0) {$X_2$};
    \node[RR] (Y) at (1,-1) {$Y$};
    \draw[->] (X1) -- (Y);
    \draw[->] (X2) -- (Y);
\end{tikzpicture}}\hfill
\subfigure[]{\label{fig:example-b}\centering
\begin{tikzpicture}[x=7mm,y=15mm]
    \node[RR] (X1) {$X_1$};
    \node[RR] (X2) at (2,0) {$X_2$};
    \node[RR] (Y) at (1,-1) {$Y$};
    \draw[->] (X1) -- (Y);
    \draw[->,dashed,gray] (X2) -- (Y) node[midway,red] {\Large $\times$};
\end{tikzpicture}}\hfill
\subfigure[]{\label{fig:example-c}\centering
\begin{tikzpicture}[x=7mm,y=15mm]
    \node[RR] (X1) {$X_1$};
    \node[RR] (X2) at (2,0) {$X_2$};
    \node[RR] (Y) at (1,-1) {$Y$};
    \draw[->] (X2) -- (Y);
    \draw[->,dashed,gray] (X1) -- (Y) node[midway,red] {\Large $\times$};
\end{tikzpicture}}\hfill
\subfigure{\label{fig:example-d}\centering
\begin{tikzpicture}[x=7mm,y=15mm]
    \node[RR] (X1) {$X_1$};
    \node[RR] (X2) at (2,0) {$X_2$};
    \node[RR] (Z) at (-.75,-.6) {$Z$};
    \node[RR] (Y) at (1,-1) {$Y$};
    \draw[->] (X1) -- (Y);
    \draw[->] (X2) -- (Y);
    \draw[->] (X1) -- (Z);
    \draw[->] (X2) -- (Z);
    \draw[->] (Z) -- (Y);
\end{tikzpicture}}\hfill
\subfigure{\label{fig:example-e}\centering
\begin{tikzpicture}[x=7mm,y=15mm]
    \node[RR] (X1) {$X_1$};
    \node[RR] (X2) at (2,0) {$X_2$};
    \node[RR] (Z) at (-.75,-.6) {$Z$};
    \node[RR] (Y) at (1,-1) {$Y$};
    \draw[->] (X1) -- (Y);
    \draw[->,dashed,gray] (X2) -- (Y) node[midway,red] {\Large $\times$};
    \draw[->] (X1) -- (Z);
    \draw[->] (X2) -- (Z);
    \draw[->] (Z) -- (Y);
\end{tikzpicture}}\hfill
\subfigure{\label{fig:example-f}\centering
\begin{tikzpicture}[x=7mm,y=15mm]
    \node[RR] (X1) {$X_1$};
    \node[RR] (X2) at (2,0) {$X_2$};
    \node[RR] (Z) at (-.75,-.6) {$Z$};
    \node[RR] (Y) at (1,-1) {$Y$};
    \draw[->,dashed,gray] (X1) -- (Y) node[midway,red] {\Large $\times$};
    \draw[->] (X2) -- (Y);
    \draw[->] (X1) -- (Z);
    \draw[->] (X2) -- (Z);
    \draw[->] (Z) -- (Y);
\end{tikzpicture}}
\caption{(a) Causal graph in \Cref{ex:CSSI1}. (b, c) CSI can represent the local independence of $Y$ and $X_2$, but not the other. In contrast, \nc{} is able to represent both.\comment{may draw a labeled DAG?} (d) Augmented causal graph with \nv{} $Z$ added in \Cref{ex:CIPV3}. (e, f) \nv{} representing \nc{}s for each context set.}
\label{fig:example}
\end{figure}
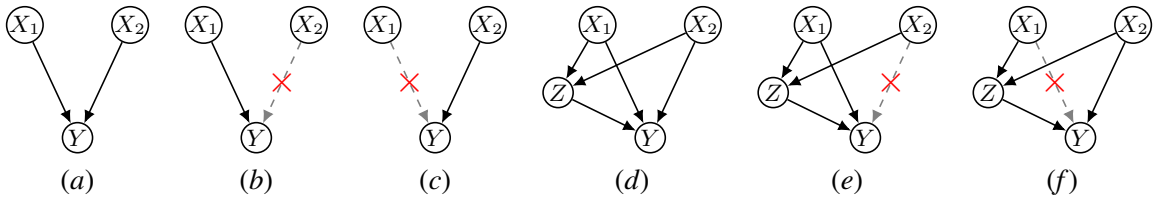

Consider a system as an example consisting of three observed variables $X_1$, $X_2$, $Y$, and an unobserved one $U$. We describe the system using a structural causal model (SCM) \citep{pearl2009causality, peters2017elements}. Let $X_1, X_2$ follow a uniform distribution between 0 and 1, and $U$ follow a standard Gaussian distribution. Further, let $Y$ be determined as $Y = X_1 + U$ if $X_1X_2 < \sfrac{1}{2}$ and $Y = X_2 + U$ otherwise. Focusing on the observable variables, while $Y$ depends on both $X_1$ and $X_2$ overall (\Cref{fig:example-a}),  
$Y$ depends only (i) on $X_1$ under a certain condition and (ii) on $X_2$, otherwise. Since $X_1<\sfrac{1}{2}$ implies the condition $X_1X_2<\sfrac{1}{2}$ and $Y$ being the function of $X_1$ and $U$, it can be represented as CSI (\Cref{fig:example-b}).
On the other hand, CSI cannot capture the other local independence where $Y$ is determined only by $X_2$ and $U$ since $Y$ and $X_1$ are always dependent given any assignment of $X_2$.

Against this background, we define \newcsi{} (\nc{}), a generalized notion of local independence where conditional independence holds in a certain \textit{context set}, a set of joint assignments of the variables. 
We characterize \newcsi{} in order to uncover \nc{}s for continuous variables, which is more challenging due to an infinite number of context sets. 
Our characterization leads to the design of \ours{} (\oursFull{}) to find the partitions of the joint outcome space where each partition set induces a \nc.
Our approach is based on approximating a conditional distribution for each partition given a subset of parents. For instance, based on the above example, our method learns two partitions of the Cartesian product of the domains of $X_1$ and $X_2$, where each partition is flexibly modeled as $\hat{P}(Y|X_1)$ and $\hat{P}(Y|X_2)$.

Our contributions are summarized as follows.
(i) We introduce the notion of \newcsi{} (\nc), a new class of local conditional independence relationships. This generalizes two well-known local independence, CSI and partial conditional independence (PCI), and permits continuous conditioned variables.
(ii) We characterize \newcsi{} by relating possible \nc{}s arisen from a given distribution. We examine a canonical representation of \nc{}s and provide a sufficient condition under which a unique \nc{}s exists. These characterizations lead to defining contextual decomposition, a way to understand an entire group of \nc{}s presenting in a given dataset.
(iii) We devise \oursfull{}, a simple and effective method for finding a contextual decomposition without directly testing for individual local independence. It utilizes an auxiliary variable (\newvariable{}) used for training of a neural network for context decomposition. Empirical evaluations on a synthetic dataset and Spriteworld demonstrate the effectiveness of the proposed method in recovering local independence relationships.

\section{Preliminaries}
\label{sec:preliminaries}
Throughout this paper, we use capital letters for random variables and small letters for their assignments. Bold letters denote the set of random variables or its assignments. Calligraphy letters can be the domain of corresponding variables or complex mathematical objects.

\subsection{Structural Causal Model}
We adopt a structural causal model (SCM) \citep{pearl2009causality} as a causal framework to understand generated data. An SCM $\mathcal{M}$ is defined as a tuple $\left< \mathbf{V}, \mathbf{U}, \mathbf{F}, P(\mathbf{U}) \right>$, where $\mathbf{V}= \{V_1, \cdots, V_d\}$ and $\mathbf{U}$ are a set of endogenous and exogenous variables, respectively. $\mathbf{F} = \{f_1, \cdots, f_d\}$ is a set of functions determining each endogenous variable, i.e., $V_j \gets f_j(Pa(j), \*U_j)$ where $Pa(j) \subseteq \*V\setminus \{V_j\}$ is a parent of $V_j$ and $\*U_j\subseteq \*U$. We may use variable $V_i$ and its index $i$ interchangeably. $\mathcal{V}_{j}$ is the outcome space of each variable $V_j$. In this work, we restrict to a Markovian model\footnote{Conditional independence is a statistical notion and would be irrelevant to whether the model is Markovian or semi-Markovian. However, Markovian models will allow clearer interpretation of context-specific independence with respect to the functional form of the target variable irrespective of regimes, i.e., observational or interventional.} and assume that an SCM satisfies structural minimality \citep{pearl2009causality, peters2017elements}, which asserts that 
$j \in Pa(i)$ if and only if there exist two values for $j$ that lead to different values for $i$, i.e., $f_i(v_j, \mathbf{v}_{Pa(i)\setminus \{j\}}, \*u_i) \neq f_i(v'_j, \mathbf{v}_{Pa(i)\setminus \{j\}}, \*u_i)$ where $v_j, v'_j \in \mathcal{V}_j$ and $\mathbf{v}_{Pa(i)\setminus \{j\}} \in \mathcal{V}_{Pa(i)\setminus \{j\}}$.
An SCM $\mathcal{M}$ induces a causal graph $\mathcal{G}=(\mathbf{V}, \mathbf{E})$, which is a directed acyclic graph (DAG), where $\*V=\{1, \ldots, d\}$ and $\*E \subseteq \*V \times \*V$ are the set of nodes and edges, respectively. If $i\in Pa(j)$ in $\mathcal{M}$, then a directed edge $(i, j)$ is in $\mathbf{E}$. 
In the language of SCM, each node $i$ corresponds to a random variable $V_i$ and each edge $(i, j)$ denotes a direct causal relationship from $V_i$ to $V_j$.

\subsection{Local Independence Relationship of Discrete Variables}

As described earlier, in many cases, the causal mechanism exhibits fine-grained independence relationships, which do not hold in general but only under certain conditions, i.e., \textit{local independence}. To begin with, we first provide the definition of context-specific independence (CSI) and partial conditional independence (PCI).
We adopt the definitions of CSI and PCI from prior works  \citep{boutilier2013contextspecific, Pensar_2014,pensar2016role} for the relationship between a variable and its parents. For the definitions of CSI and PCI, variables are assumed to be discrete. Let $\rmX$ be a non-empty set of parents of $Y$ and $\rmX_{A}, \rmX_{B}$ be non-empty partitions of $\*X$, i.e., $\*X$ is disjoint union of $\*X_A$ and $\*X_B$.
\begin{definition}[Context-Specific Independence (CSI)]
\label[definition]{def:CSI}
$Y$ is said to be \emph{contextually independent} of $\rmX_{B}$ given the context $\rmX_{A}=\rvx_{A}$ if $P\left(y \mid \rvx_{A}, \rvx_{B}\right)=P\left(y \mid \rvx_{A}\right)$, holds for all $y\in \mathcal{Y}$ and $\rvx_{B} \in \mathcal{X}_{B}$ whenever $P\left(\rvx_{A}, \rvx_{B}\right)>0$. This will be denoted by $Y \Perp \rmX_{B} \mid \rmX_{A} = \rvx_{A}$.
\end{definition}
\begin{definition}[Partial Conditional Independence (PCI)]
\label[definition]{def:PCI}
$Y$ is said to be \emph{partially conditionally independent} of $\rmX_{B}$ in the domain $\mathcal{D}_{B}\subseteq \mathcal{X}_{B}$ given the context $\rmX_{A}=\rvx_{A}$ if
    $P\left(y \mid \rvx_{A}, \rvx_{B}\right)
    ={P(y \mid \rvx_{A}, \rvx'_{B})}$,
holds for all $y\in \mathcal{Y}$ and $\rvx_{B}, \rvx_{B}' \in \mathcal{D}_{B}$ whenever $P\left(\rvx_{A}, \rvx_{B}\right)>0$ and $P(\rvx_{A}, \rvx'_{B})>0$. This will be denoted by $Y \Perp \rmX_{B} \mid \mathcal{D}_{B}, \rmX_{A} = \rvx_{A}$.
\end{definition}
PCI is a more fine-grained notion than CSI and, thus, PCI does not imply CSI: $P(y\mid \*x_A, \*x_B) = {P(y\mid \*x_A, \*x_B')}$ for all $\rvx_{B}, \rvx_{B}' \in \mathcal{D}_{B}$ is not equivalent to $P(y\mid \*x_A, \*x_B) = P(y\mid \*x_A)$.
Further discussions of the related works on local independence relationships are provided in \Cref{sec:related_work,sec:appendix_ci}.

\section{Compactly Representing Local Independence of Continuous Variables}
\label{sec:CSSI}

\begin{wrapfigure}{r}{0.25\linewidth}
\centering
\vspace{-10pt}
\includegraphics[width=.95\linewidth, clip,trim=0 0 6mm 2mm]{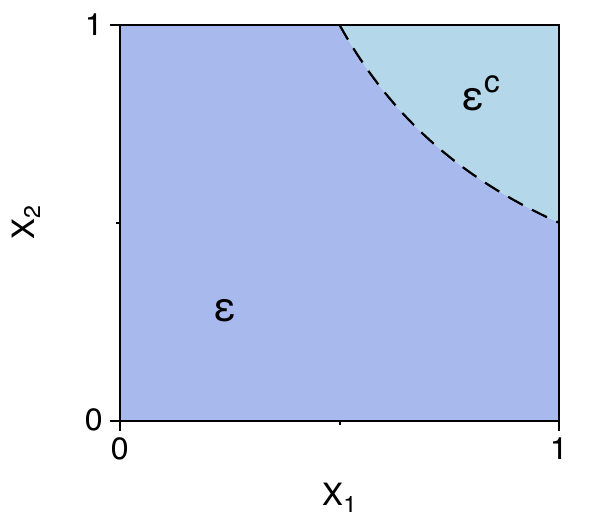}
\caption{\Cref{ex:CSSI1}.}
\label{fig:ex1}
\end{wrapfigure}
Our goal is to find such local independence relationships in a system that may contain continuous variables. The information about local independence relationships of continuous variables can also be leveraged in various tasks as demonstrated in the case of discrete variables. 
Given that a naive generalization of CSI or PCI to continuous variables seems practically implausible, we first define an alternative notion of local independence, which we call \newcsi{} (\nc{}).  
Then, we characterize \nc{} and describe how it can be equipped within the framework of SCM.
All omitted proofs can be found in the Appendix.

\subsection{Local Independence Relationship over Continuous Variables}

Here, we focus on a causal mechanism of a target variable $Y$ and its parents $\*{X}=\{X_1, \cdots, X_d\}$, i.e., SCM $\mathcal{M} = \left< \*{V}, \*{U}, \*{F}, P(\*{U}) \right>$ where $\{X_1, \cdots, X_d, Y\} \subseteq \*{V}$ is a set of continuous random variables and $Y = f(\*{X}, U)$. 
We denote $\rmX_{A} = \{X_i\mid i\in A\}$ where $A\subseteq Pa(Y)= \{1, \cdots, d\}$.
Similarly, $\rmX_{A^c} = \{X_i\mid i\in A^c\}$ where $A^c = Pa(Y)\setminus A$.
Throughout the paper, we assume strictly positive densities.
We first define a \newcsi{} (\nc). 
\begin{definition}[\NewCsi{}]
\label[definition]{def:CSSI}
Let $\*{X}=\{X_1, \cdots, X_d\}$ be a non-empty set of the parents of $Y$ in a causal graph,
and $\gE\subseteq \gX$ be an event with a positive probability. $\gE$ is said to be a {\bf context set} which induces {\bf \newcsi{} (\nc)} of $\*{X}_{A^c}$ from $Y$ if $p\left(y \mid \*{x}_{A^c}, \*{x}_{A}\right)=p\left(y \mid \*{x}'_{A^c}, \*{x}_{A}\right)$ holds for every $\left(\*{x}_{A^c}, \*{x}_{A}\right), \left(\*{x}'_{A^c}, \*{x}_{A}\right) \in \gE$. This will be denoted by $Y \Perp \*{X}_{A^c} \mid \*{X}_{A}, \gE$.
\end{definition}

\nc{} extends CSI and PCI in the sense that the independence holds in a set of conditioned values in \nc{} since they consider a point-based condition. 
We revisit the earlier example to illustrate how CSI, PCI, and \nc{} (\Cref{def:CSI,def:PCI,def:CSSI}) represent the local independence relationships in the system (insufficiency results for CSI and PCI in representing \Cref{ex:CSSI1} is depicted in \Cref{sec:appendix_ci}.)

\begin{example}
\label{ex:CSSI1}
Let $X_1, X_2$ be a uniform random variable defined on $[0, 1]$, s.t. $X_1, X_2 \sim \textrm{Unif }[0, 1]$ and $U$ be an exogenous variable. Let $Y$ be: 
\begin{align*}
Y = 
\begin{cases}
X_1 + U & \text {if }\quad \,\,\,X_1X_2 < \sfrac{1}{2}, \\
X_2 + U & \text {if }\quad \,\,\,X_1X_2 \geq \sfrac{1}{2}.
\end{cases}
\end{align*}
Let $\gE=\{(x_1, x_2) \mid x_1 x_2 < \sfrac{1}{2} \}$. Then, $Y \Perp X_2 \mid X_1, \gE$ and $Y \Perp X_1 \mid X_2, \gE^c$ hold.
On the other hand, $Y \not\Perp X_1 \mid X_2=c$ for any $c$. 
\end{example}

As described earlier, we focus on the discovery of local independence (i.e., \nc{}) of continuous variables and henceforth assume variables are continuous. In a case where the variables are discrete, \nc{} can be discovered by first (i) discovering PCI relationships and then (ii) integrating them (see \Cref{sec:appendix_ci} for the detail). 
We now introduce important properties of CSSI. 
\begin{restatable}[{CSSI Entailment}]{re-proposition}{regularcssi}
\label[proposition]{prop:regularcssi}
 Suppose a \nc{} relationship $Y\Perp \*X_{A^c} \mid \*X_{A}, \gE$ holds. Then, the following \nc{} relationships also hold:
 
 \hfill(i) $Y\Perp \*X_{B^c} \mid \*X_{B}, \gE\quad$ for any $B\supseteq A$, \hfill
 (ii) $Y\Perp \*X_{A^c} \mid \*X_{A}, \gF\quad$ for any $\gF \subseteq \gE$.\hfill\null
\end{restatable}
Due to such implications, we are interested in \nc{} with \textit{minimal} conditioned variables.
\begin{definition}
A \nc{} of the form $Y\Perp \*X_{A^c} \mid \*X_{A}, \gE$ is said to be \textbf{regular} if there does not exists any $B\subsetneq A$ such that $Y\Perp \*X_{B^c} \mid \*X_{B}, \gE$ holds. 
Then, we say $\*X_{A}$ to be a \textbf{local parent set} of $Y$ in $\gE$.\footnote{We use $\*{pa}^\gE$ and $Pa^\gE(Y)$ to denote the local parent set of $Y$ in $\gE$, i.e., $\*{pa}^\gE=\rmX_{A}$ and $Pa^\gE(Y) = A$.}
\end{definition}

\begin{wrapfigure}[10]{r}{0.25\linewidth}
\centering
\vspace{-10pt}
\includegraphics[width=.95\linewidth]{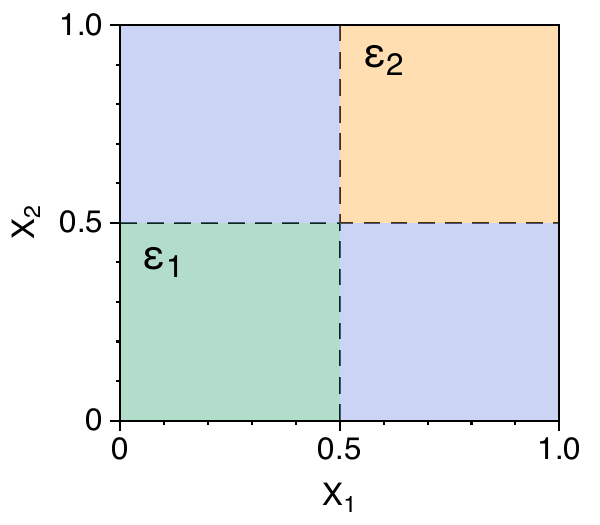}
\caption{\Cref{ex:regularcssi}.}
\label{fig:ex2}
\end{wrapfigure}
Among the \nc{}s with the same context set, the regular \nc{}s are the most \textit{informative}, in the sense that the set of the conditioned variables cannot be further reduced. 
We say a \nc{} is \textbf{trivial} if $A=Pa(Y)$ (i.e., $A^c=\emptyset$ and $\rmX=\rmX_{A}$), since it trivially holds for any $\gE\subseteq \gX$.
A trivial \nc{} on $\gE=\gX$ is indeed regular.
In general, regular \nc{} may not be unique on the given context set. We show the non-uniqueness of local parent sets by an example below. 
\begin{example}
\label{ex:regularcssi}
Let $X_1, X_2, X_3 \sim \textrm{Unif }[0, 1]$ and $U$ be an exogenous variable. 
Let $Y$ be: 
\begin{align*}
Y = 
\begin{cases}
X_3 + U & \text {if}\quad X_1, X_2 < \sfrac{1}{2}, \\
X_3 + 2\cdot U & \text {if}\quad X_1, X_2 \geq \sfrac{1}{2}, \\
X_1 + X_2 + X_3 + U & \text {otherwise}.
\end{cases}
\end{align*}
Let $\gE_1=\{\*x \mid x_1, x_2 < \sfrac{1}{2} \}$, $\gE_2=\{\*x \mid x_1, x_2 \geq \sfrac{1}{2} \}$.
Then, regular \nc{}s 
$Y \Perp X_{12} \mid X_{3}, \gE_1$ and 
$Y \Perp X_{12} \mid X_{3}, \gE_2$ hold.
Also, the following regular \nc{}s hold on $\gE_1\cup\gE_2$:
\begin{align*}
Y \Perp X_1 \mid X_{23}, \gE_1\cup\gE_2 \qquad
\text{and} \qquad
Y \Perp X_2 \mid X_{13}, \gE_1\cup\gE_2.
\end{align*}
\end{example}

In the example above, there exist two distinct regular \nc{}s on $\gE_1\cup\gE_2$. 
If the local parent set is not unique on the given context set, discovering the local parent set would yield an inconsistent result. Thus, we explore a sufficient condition that guarantees the uniqueness of the local parent set. We now provide an intersection property of \nc{}, which will be used to derive a sufficient condition.
\begin{restatable}[Intersection Property of \nc{}]{re-proposition}{convexintersection}
\label[proposition]{prop:convexintersection}
Suppose $\gE\subseteq\gX$ is convex. 
If \nc{} relationships $Y\Perp \*X_{A^c} \mid \*X_{A}, \gE$ and $Y\Perp \*X_{B^c} \mid \*X_{B}, \gE$ hold, then $Y\Perp \*X_{(A\cap B)^c} \mid \*X_{A\cap B}, \gE$ hold.
\end{restatable}

In \Cref{ex:regularcssi}, we cannot derive $Y \Perp X_{12} \mid X_{3}, \gE_1\cup\gE_2$ from the two regular CSSIs on $\gE_1\cup\gE_2$ as $\gE_1\cup\gE_2$ is not convex.
We now introduce a sufficient condition that guarantees the uniqueness of regular \nc{}s given a context set $\gE$. 
\begin{restatable}[Uniqueness of Local Parent Set]{theorem}{convexuniqueregular}
\label[theorem]{thm:convexuniqueregular}
Suppose $\gE\subseteq\gX$ is convex. 
There exists a unique $A\subseteq Pa(Y)$ such that the regular \nc{} relationship $Y\Perp \*X_{A^c} \mid \*X_{A}, \gE$ holds.
\end{restatable}
\begin{proof}
Suppose regular CSSIs $Y\Perp \*X_{A^c} \mid \*X_{A}, \gE$ and $Y\Perp \*X_{B^c} \mid \*X_{B}, \gE$ hold for some $A$ and $B$. 
By \Cref{prop:convexintersection} and the convexity of $\gE$, $(Y\Perp \*X_{(A\cap B)^c} \mid \*X_{A\cap B}, \gE)$ holds true. Due to the minimality (i.e., regular), $A\cap B$ cannot be a proper subset of $A$ nor $B$. Hence, $A=B=A\cap B$.
\end{proof}
This implies that one may devise a conditional independence test or causal discovery algorithm 
to find the (unique) local parent set on the convex subset of data (e.g., a rectangular $\gE = [a_1, b_1]\times \cdots \times [a_d, b_d] \in \mathbb{R}^d$ or a ball $\gE = \{\*{x}\mid \|\*{x} - \*{x}_0\|_2 \leq r\}$). 
If the subset of data is non-convex, the local parent set \emph{may} not be unique,\footnote{Note that this does not imply that local parent set is \emph{always} not unique on the non-convex subset of data, e.g., $Y \Perp X_2 \mid X_1, \gE$ is the unique regular \nc{} on the non-convex set $\gE$ in \Cref{ex:CSSI1}.} and thus the results may not be valid without any additional assumptions.
We provide a more general condition for \Cref{prop:convexintersection} and \Cref{thm:convexuniqueregular} in \Cref{sec:appendix_coordinate}.

\section{Discovering Local Independence Relationships by Learning the Partition}
\label{sec:CIPV}
We investigate the representation and discovery of \emph{multiple} \nc{}s in a system.
To compactly represent \nc{}s, we introduce a decomposition of the parents' outcome space, where each partition corresponds to a \nc{}-inducing context set.
Then, we transform the finding of such decomposition as a learning problem involving an auxiliary variable representing the decomposition. 
Finally, we develop a neural approach to learning the decomposition via the auxiliary variable.

\subsection{Representing \nc{}s in a System as a Partition}
We have seen through examples that a system may exhibit multiple \nc{}s.
We define a canonical notion of \nc{} in order to examine a compact representation of \nc{}s.

\begin{definition}
A \nc{} relationship $Y\Perp \*X_{A^c} \mid \*X_{A}, \gE$ is \textbf{canonical} if there does not exists any $\gF \subseteq \gE$ such that $Y\Perp \*X_{B^c} \mid \*X_{B}, \gF$ and $B\subsetneq A$.
\end{definition}

By definition, the canonicality of a CSSI implies its regularity.
Canonical \nc{}s are the most {fine-grained} ones since any subset of the context set also entails the same regular \nc{} relationship. 
In \Cref{ex:regularcssi}, \nc{} relationship $Y \Perp X_1 \mid X_{23}, \gE_1\cup\gE_2$ is regular but not canonical since $Y \Perp X_{12} \mid X_{3}, \gE_1$.
If a trivial \nc{} on $\gE=\gX$
is canonical, a system does not exhibit any \nc{} relationship (e.g., \Cref{ex:existence} in \Cref{sec:appendix_examples}).
We now define a compact representation of multiple \nc{}s in a system.

\begin{definition}[\CtxDec{}]
\label[definition]{definition:cd}
Let $\{\gE_k\}_{k=0}^N$ be a partition of $\gX$.
$\{(\gE_k, {A_k})\}_{k=0}^N$ is a \textbf{\ctxdec{} (\cd{})} if regular \nc{} $Y\Perp \*{X}_{A_k^c} \mid \*{X}_{A_k}, \gE_k$ holds for all $k$  
where $A_0=Pa(Y)$ and $A_k\subsetneq Pa(Y)$ for all $k\geq 1$.\footnote{While the context sets have positive probability by its definition, we allow the case of $p(\gE_0)=0$.}
We say a \cd{} $\{(\gE_k, A_k)\}_{k=0}^N$ is \textbf{canonical} if $Y\Perp \*X_{A_k^c} \mid \*X_{A_k}, \gE_k$ is canonical for all $k$, and is \textbf{distinctive} if $A_i \neq A_j$ for all $i\neq j$.
\end{definition}

$\gE_0$ is a set to cover a remaining subset of $\gX$ excluding non-trivial context sets. $(\gE_0, A_0)$ corresponds to a trivial \nc{}.
For any causal system, $\{(\gX, Pa(Y))\}$ is a \cd{}, which we call \textbf{trivial \cd{}}. 
When a system does not exhibit any \nc{} relationship, a trivial \cd{} is the only existing \cd{}.
In \Cref{ex:regularcssi}, both $\{(\gE_0, \rmX_{123}), (\gE_1, \rmX_3), (\gE_2, \rmX_3)\}$ and $\{(\gE_0, \rmX_{123}), (\gE_1 \cup \gE_2, \rmX_{23})\}$ are \cd{}s where $\gE_0 = (\gE_1 \cup \gE_2)^c$.
The former is canonical, but the latter is not.
\Ctxdec{} can be viewed as the discretization of the joint outcome space, in contrast to discretizing each variable which is a commonly used strategy to handle continuous variables \citep{nyman2017stratified}.
As a canonical \nc{} is a fine-grained notion of \nc{} relationship, a canonical \cd{} can also be viewed as a fine-grained \cd{}. 
While a canonical \cd{} may not be unique (e.g., \Cref{ex:canonical_cd} in \Cref{sec:appendix_examples}), the following theorem characterizes the shared property of the canonical \cd{}s in a system.
\begin{restatable}{theorem}{cdatomic}
\label[theorem]{theorem:cdatomic}
Let $\{(\gE_i, A_i)\}_{i=0}^N$ and $\{(\gF_j, B_j)\}_{j=0}^M$ be canonical \cd{}s where $\gE_i$ and $\gF_j$ are open sets for all $i, j\geq 1$.
Then, the following holds: (i) if $p(\gE_i \cap \gF_j)>0$ then $A_i=B_j$, and (ii) for any $C\subseteq Pa(Y)$, $p(\gE[C] \bigtriangleup \gF[C])=0$\footnote{$\bigtriangleup$ is a symmetric difference, i.e., $A\bigtriangleup B = (A\setminus B) \cup (B\setminus A)$} where $\gE[C]=\bigcup_{A_i=C} \gE_i$ and $\gF[C]={\bigcup}_{B_j=C} \gF_j$.
\end{restatable}

Roughly, any intersecting context sets (i.e., $p(\gE_i \cap \gF_j)>0$) from different canonical \cd{}s share the same local parent set (i.e., $A_i=B_j$).
Also, for any $C\subseteq Pa(Y)$, the union of the context sets having the same local parent set $C$ is identical for any canonical \nc{} in a system (i.e., $p(\gE[C] \bigtriangleup \gF[C])=0$).
We provide an example of canonical \cd{}s to elaborate \Cref{theorem:cdatomic} in \Cref{ex:canonical_cd} (\Cref{sec:appendix_examples}).
An interesting property of \cd{}s is that the union of the context sets having the same local parents set may no longer entail the same one, i.e., $\*{pa}^{\gE_1 \cup \gE_2} \neq \*{pa}^{\gE_1} \cup \*{pa}^{\gE_2}$, in general as shown in \Cref{ex:regularcssi}.
For the case of distinctive canonical \cd{}s, we have stronger uniqueness which directly follows from \Cref{theorem:cdatomic}.
\begin{restatable}[Uniqueness of Distinctive Canonical \cd{}]{re-corollary}{cdatomicnonidentical}
\label[re-corollary]{corollary:cdatomicnonidentical}
Let $\{(\gE_i, A_i)\}_{i=0}^N$ and $\{(\gF_j, B_j)\}_{j=0}^M$ be distinctive canonical \cd{}s where $\gE_i$ and $\gF_j$ are open sets for all $i, j\geq 1$.
Then, the following hold: (i) $N=M$, and (ii) if  $p(\gE_i \cap \gF_j)>0$ then $A_i=B_j$ and $p(\gE_i\bigtriangleup \gF_j)=0$.
\end{restatable}

In words, distinctive canonical CD is \emph{unique} since any intersecting context sets (i.e., $p(\gE_i \cap \gF_j)>0$) from different distinctive canonical \cd{}s share the same local parent set (i.e., $A_i=B_j$), and their difference is negligible (i.e., $p(\gE_i\bigtriangleup \gF_j)=0$).

\subsection{Representing \CtxDec{} with an Auxiliary Variable}
We now introduce a \newvariable{}, which will serve as a mapping between $\gX$ and an arbitrary contextual decomposition. 
This variable will be useful in transforming the task of testing for CSSI to the task of learning such decomposition.
We first show that a \cd{} can be represented by introducing an auxiliary variable.

\begin{restatable}[Expressing \nc{} as CSI with Auxiliary Variable]{re-proposition}{cdandcsi}
\label[proposition]{prop:cdandcsi}
Let $\{\gE_k\}_{k=0}^N$ be a partition of $\mathcal{X}$.
$\{(\gE_k, A_k)\}_{k=0}^N$ is a contextual decomposition
if and only if $Y\Perp \*X_{A_k^c} \mid \*X_{A_k}, Z=z_k$ holds for all $k\geq 1$, where $Z$ is the deterministic random variable defined as $Z=z_k$ if $\*X\in\gE_k$ for all $k$.
\end{restatable}

In \Cref{ex:CSSI1}, we can introduce $Z$ so that $Y \Perp X_2 \mid X_1, Z=z_1$ and $Y \Perp X_1 \mid X_2, Z=z_2$ where $Z=z_1$ if $\rmX\in \gE$ and $z_2$ otherwise. 
We now formally define a \newvariable{}.

\begin{definition}
\label[definition]{definition:piv}
Let $\{\gE_k\}_{k=0}^N$ be a partition of $\mathcal{X}$ and a random variable $Z$ be defined as $Z=z_k$ if $\*X\in\gE_k$ for all $0\leq k \leq N$. 
Variable $Z$ is a \textbf{\newvariable{} (\nv{})} if for all $k\geq 1$ there exists some $A_k\subsetneq Pa(Y)$ such that $Y\Perp \*X_{A_k^c} \mid \*X_{A_k}, Z=z_k$.
\end{definition}

PIV can be viewed as a particular type of sufficient set of statistics.\footnote{\cite{chicharro2020causal} utilized a sufficient set of statistics with a focus on discovering a causal structure. 
However, we provided the characterization and representation of local independence and its fundamental properties.
Although the concept of a sufficient set of statistics and related rules is valid for continuous variables, their information bottleneck approach is {restricted to the discrete variables}.} 
\Cref{prop:cdandcsi} implies that a \nv{} represents a corresponding \cd{}.
With \nv{} $Z$, each \nc{} relationship $Y\Perp \*X_{A_k^c} \mid \*X_{A_k}, \gE_k$ is equivalently expressed as $Y\Perp \*X_{A_k^c} \mid \*X_{A_k}, Z=z_k$.
Thus, for $\rvx\in \gE_k$, 
\begin{equation}
p(y\mid \rvx) = p(y\mid \rvx, z_k) = p(y\mid \rvx_{A_k^c}, \rvx_{A_k}, z_k) = p(y\mid \rvx_{A_k}, z_k),
\end{equation}
where the first equality holds by definition, and the last equality holds since \nv{} entails CSI relationships.
Therefore some of the parent variables (i.e., $\rmX_{A_k^c}$) would be ignored for modeling the conditional distribution of $Y$ given $\gE_k$.
In contrast to CSI, which uses the value of a proper subset of $Pa(Y)$ ignoring the rest, the value of $Z$ is determined by the whole $\rmX$, in general, and a subset of $Pa(Y)$ to be ignored (i.e., $\rmX_{A_k^c}$) could involve in determining the value of $Z$.

\subsection{\oursFull}
As described earlier, conditional independence tests  can be used to discover the \nc{} relationship on a particular subset of data. Further, assuming strictly positive densities, testing on a convex subset of data is sufficient to obtain a unique local parent set without any additional assumptions. 
However, it is generally infeasible to test on every possible subset to discover multiple \nc{}s in a given system.

Against this background, we propose \textbf{\oursFull{} (\ours)}, a neural approach to recovering \emph{distinct} contextual decomposition from given data $P(\*V)$ via learning \nv{} $Z$.
Recalling $Y\Perp \*X_{A_k^c} \mid \*X_{A_k}, Z=z_k$, we let $\gZ\subseteq \{0,1\}^d$ so that the local parent set $A_k$ corresponds to the set of the indices of the nonzero element of $z_k$, e.g., $Y \Perp X_2 \mid X_{13}, Z=(1, 0, 1)$.\footnote{We assume that $Pa(Y)=\{1, \cdots, d\}$ is identifiable and correctly discovered, e.g., by some causal discovery methods.}
With \nv{} $Z$, we can write the conditional density as:
\begin{align}
\textstyle{
p(y\mid \rvx) = \sum_z p(y\mid \rvx, z) p(z\mid \rvx) 
= \sum_z p(y\mid \rvx_{A_z}, z) p(z\mid \rvx),}
\end{align}
where $A_z$ is $A_k$ corresponding to $z_k$ and $p(z\mid \rvx) = \delta(z=z_i)$ if $\rvx\in \gE_i$. Our method models $p(y\mid \rvx, z)$ and $p(z\mid \rvx)$.

\paragraph{Modeling $p(y\mid \rvx, z)$ and $p(z\mid \rvx)$.}
Since $p(y\mid \rvx, z) = p(y\mid \rvx_{A_z}, z)$, some of the parent variables (i.e., $\rvx_{A_z^c}$) are redundant for modeling a conditional distribution.
Therefore, our method approximates the conditional distribution $p(y\mid \rvx, z)$ as $\hat{p}(y\mid \rvx, z)$ and let a neural network $f_\theta$
takes $(\*{x}\odot z, z)$ as an input where $\odot$ denotes an element-wise product and outputs the parameters of the estimator $\hat{p}(y\mid \*{x}, z)$.\footnote{In our experiments, we model $\hat{p}$ as the Gaussian and let $f_\theta$ outputs the parameters of the Gaussian distribution.}
For the input of $f_\theta$, we simply concatenate $\rvx\odot z$ and $z$, e.g., if $z=(1,0,1)$ then $f_\theta$ takes $(\rvx\odot z, z) = (x_1, 0, x_3, 1, 0, 1)$ as an input.
We emphasize that labeling (i.e., concatenation of $z$) is essential since $\rvx_{A_k^c}$ cannot be ignored without conditioning on $z$, i.e., $p(y\mid \rvx) = p(y\mid \rvx, z_k) = p(y\mid \rvx_{A_k}, z_k)$ holds for $\rvx\in \gE_k$ but $p(y\mid \rvx) \neq p(y\mid \rvx_{A_k})$ and $p(y\mid \rvx, z_k) \neq p(y\mid \rvx_{A_k})$ in general.
Moreover, without labeling, it is unable to distinguish whether the zero entries of input are either masked or its assignment being zero.
We also model the conditional distribution $p(z\mid \rvx)$ as $\hat{p}(z\mid \rvx)$, and let a neural network $g_\phi$ 
takes $\rvx$ as input and outputs the parameters of Bernoulli variables. 
That is, $g_\phi (\rvx) = (\pi_1, \cdots, \pi_d)$ and $z_{(k)}| \rvx \sim \text{Bernoulli}(\pi_k)$ for all $k$ where $z=(z_{(1)}, \cdots, z_{(d)})\in \{0, 1\}^d$.
We provide further discussions of the related works on neural network based causal discovery methods in \Cref{sec:related_work}.

\paragraph{Training objective.}
Our model becomes $p(y\mid \rvx; \theta, \phi) = {\sum_z \hat{p}(y\mid \rvx, z) \hat{p}(z\mid \rvx)}
= {\mathbb{E}_{\hat{p}(z\mid \rvx)}\hat{p}(y\mid \rvx, z)}$,
and its maximum likelihood estimation is as follows: 
\begin{equation}
\label{eq:objective_mle}
\sup_{\theta, \phi} \, \mathbb{E}_{p(\rvx, y)}
\big[\log \, \mathbb{E}_{\hat{p}(z\mid\rvx)} \,\hat{p}(y\mid \rvx, z)\big].
\end{equation}
For the inner expectation, we use the Monte Carlo approximation: 
$\frac{1}{N}\sum_{i=1}^N \hat{p}(y\mid \rvx, z^{(i)})$, where $z^{(i)}\sim \hat{p}(z\mid \rvx) $.
We also use Gumbel-Softmax estimator  \citep{jang2016categorical,maddison2016concrete} to enable the learning with neural networks. 
The final training objective is as follows:
\begin{equation}
\label{eq:objective_final}
\mathcal{L}(\theta, \phi;\mathcal{D}) 
= {\sum}_{(\*{x}, y) \in \mathcal{D}}
\big(\log \frac{1}{N}\sum_{i=1}^N \hat{p}(y\mid \rvx, z^{(i)})\big),
\end{equation}
where $z^{(i)}$ is sampled from $\hat{p}(z\mid \rvx)$  and $\mathcal{D}$ is a dataset of samples, which could be a minibatch in practice, from the joint distribution entailed by an underlying SCM.
Further details of the implementation of \ours{} are provided in \Cref{sec:appendix_implementation}.

\section{Empirical Evaluation}
\label{sec:experiment}

In this section, we evaluate our proposed method to discover \ctxdec{} and \nc{}s.
We first consider causal systems with 
different types of functional model and varying complexity of partition (\Cref{sec:experiment_synthetic}), and then a more complex system reflecting the real-world physical dynamics (\Cref{sec:experiment_spriteworld}).

We compare our method with attention-based methods \citep{pitis2020counterfactual} which aims to discover the local structure in a causal system of continuous variables, by computing the attention weights of the input $\rvx$.
It first trains a transformer \citep{vaswani2017attention} to model the conditional distribution $p(y\mid \rvx)$.
Then, it uses the attention weights $(a_1, \cdots, a_d)$ of the input $\rvx$ to discover the local independence relationship. 
Intuitively, a low attention score $a_j$ implies that the input variable $X_j$ has a weak influence on the target variable $Y$.
We also compare with a mixture model, which applies a single attention layer on top of the multiple NNs.
Local independence is discovered by first computing the approximation of the Jacobian of each NNs and then applying the weighted sum with the learned attention score. 
For the experiments on the synthetic dataset, we let the number of the NNs be equal to the ground truth number of partition sets (i.e., oracle modeling).

For the implementation of $f_{\theta}(\rvx\odot z, z)$, which outputs the parameters of the Gaussian distribution $\hat{p}(y\mid \rvx, z)$, and $g_\phi(\rvx)$, which outputs the parameters of the Bernoulli distributions, we used MLPs with 3 hidden layers and hidden units of 128.
For all experiments, we set the batch size to 1000 and trained our method and the baselines for 100 epochs.
All of our experiments were conducted with 3 different seeds, and a shaded area in a ROC plot represents a standard deviation.
A detailed description of the baseline methods and further implementation details are provided in \Cref{sec:appendix_implementation}.\footnote{Code available at: \href{https://github.com/iwhwang/NCD}{https://github.com/iwhwang/NCD}.}

\begin{figure*}[t]
\includegraphics[width=0.24\textwidth]{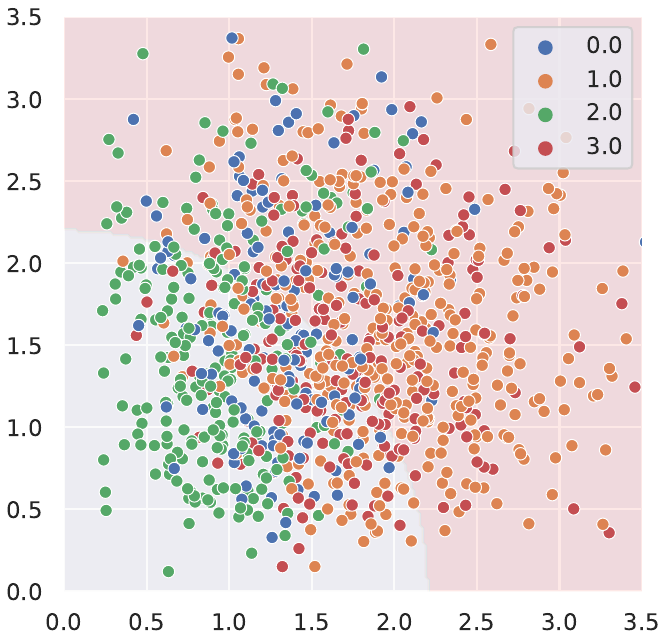}
\hfill
\includegraphics[width=0.24\textwidth]{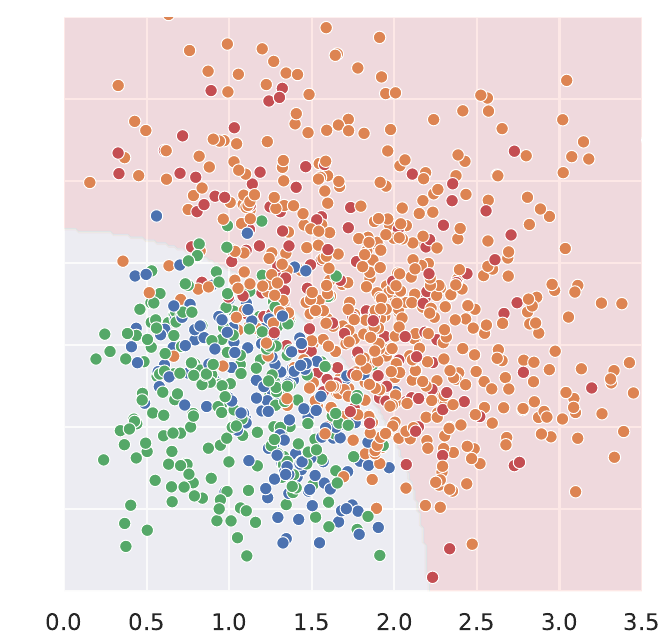}
\hfill
\includegraphics[width=0.24\textwidth]{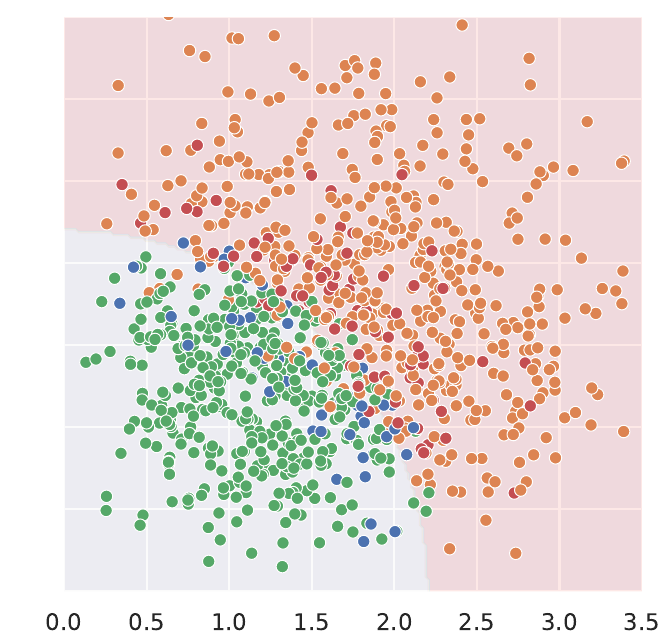}
\hfill
\includegraphics[width=0.24\textwidth]{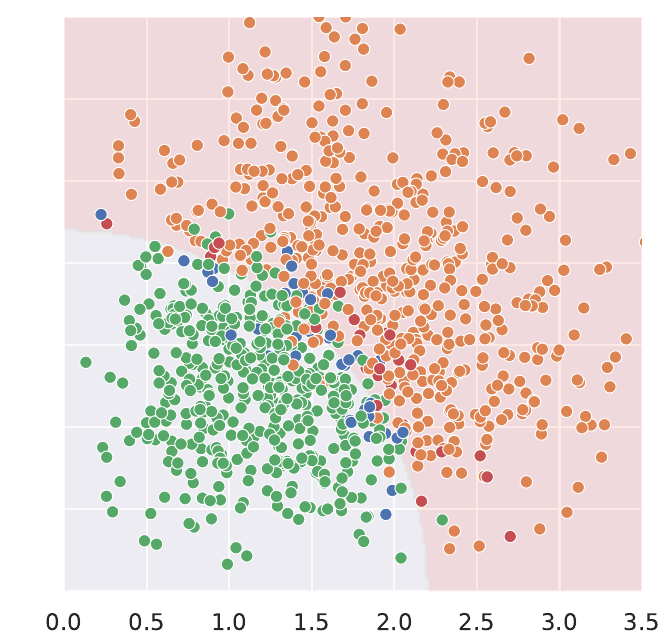}
\caption{Learned decision boundaries (partition) for epochs 5, 10, 15, and 20. Labels 0, 1, 2, and 3 denote $z=(0,0), (0,1), (1,0)$, and $(1,1)$, respectively.}
\label{fig:boundary}
\end{figure*}
\index{figure}

\subsection{Synthetic Data}
\label{sec:experiment_synthetic}
We consider causal systems where distinctive \cd{} exists.
We generate synthetic datasets consisting of the samples $(\rmX, Y)$ where $\rmX=\{X_1, \cdots, X_d\}$ is the parent variables of $Y$. 
We assume the parent variables are independent and follow the unit normal.
The data generating process is as follows:
\begin{equation}
\label{eq:atomic_system}
Y = f_i(\*{X}_{A_i}, U) \quad \text {if }\quad \*{x}\in\gE_i, \quad \text{for } \quad 0\leq i\leq N,
\end{equation}
where $U$ is an exogenous variable, $\{\gE_k\}_{k=0}^N$ is a partition of $\mathcal{X}$, $A_i$ is distinctive to each other (i.e., $A_i\neq A_j$ for all $i\neq j$).
In our experiments, $d=9$ and the exogenous variable follows the unit normal.
We evaluate our method with varying types of functions $f_k$, local parents sets $A_k$, and partition sets $\gE_k$. 
For the \textbf{functions} $\{f_k\}$, we consider nonlinear functions with (i) additive and (ii) non-additive noise models.
We use randomly initialized neural networks for the nonlinear functions, i.e., we do not assume any specific family of distributions for $f_k$.
For the partition $\{\gE_k\}$, we consider the cases where the \textbf{boundaries} of the partition sets are (i) linear and (ii) nonlinear.
In the case of linear boundaries, partition sets are defined by a linear function $h$, i.e., $x\in \gE_k$ if $h(\rmX_{A_k}) = \text{max}(h(\rmX_{A_0}), \cdots, h(\rmX_{A_N}))$.
For the nonlinear boundaries, we further control the complexity of the partition by considering both cases when nonlinear boundaries are determined by the norm of $\rvx$ and some nonlinear function $h$.
For the \textbf{local parent sets} $\{\rmX_{A_0}, \rmX_{A_1}, \rmX_{A_2}\}$, we consider two cases: (i) uniformly distributed as $\{\rmX_{123}, \rmX_{456}, \rmX_{789}\}$ and (ii) non-uniformly distributed as $\{\rmX, \rmX_{123}, \rmX_{456789}\}$.
In our experiments, we use different configurations of functions, local parent sets, and boundaries.
For the main experiments, we consider the case of nonlinear functions with a non-additive noise model and report the results under the additive noise model in the appendix.
We provide the details of the setups and implementations in \Cref{sec:appendix_synthetic}.

\paragraph{Experimental results.}
\begin{figure*}[t!]
\centering\includegraphics[height=0.25\textwidth]{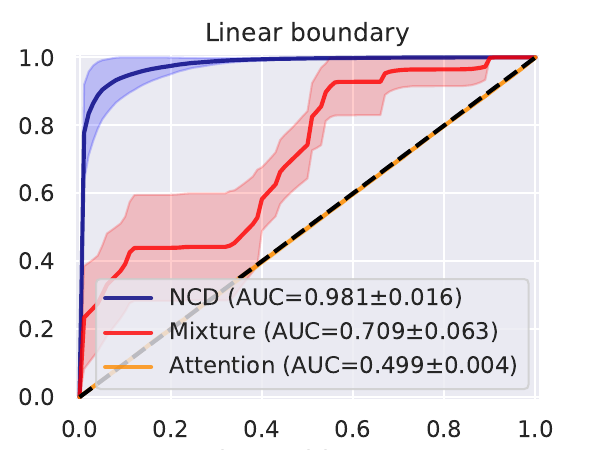}%
\hfill\includegraphics[height=0.25\textwidth]{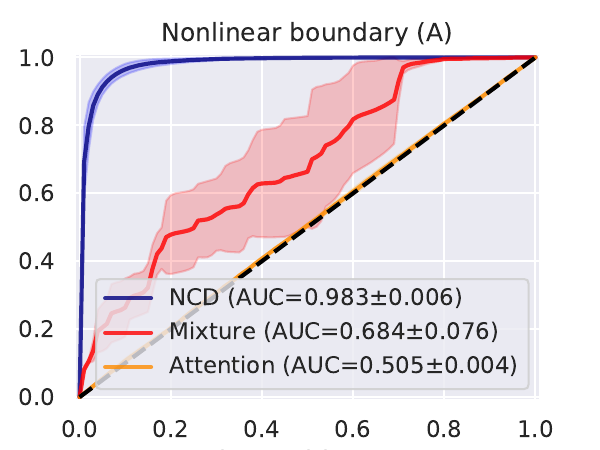}%
\hfill\includegraphics[height=0.25\textwidth]{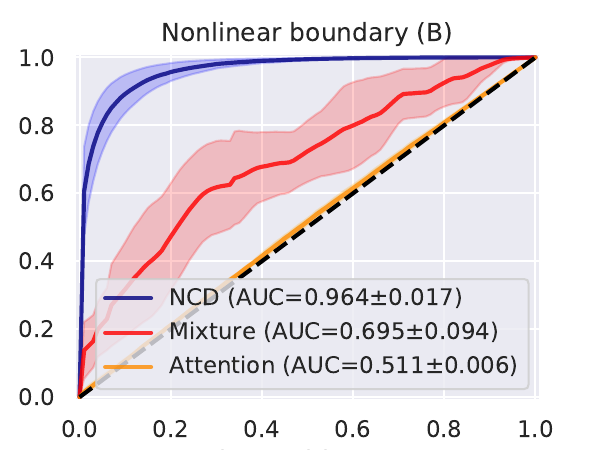}\smallskip

\includegraphics[height=0.25\textwidth]{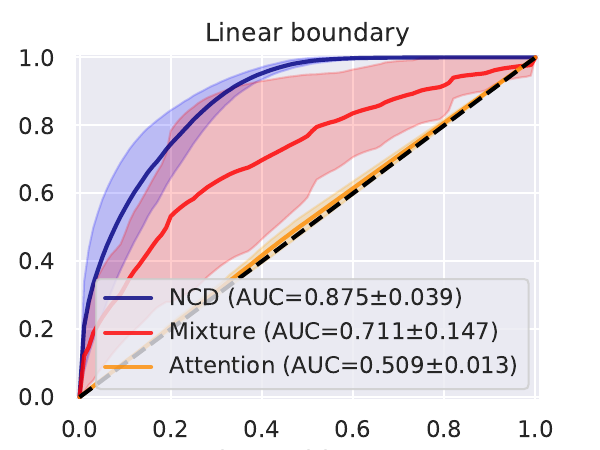}%
\hfill\includegraphics[height=0.25\textwidth]{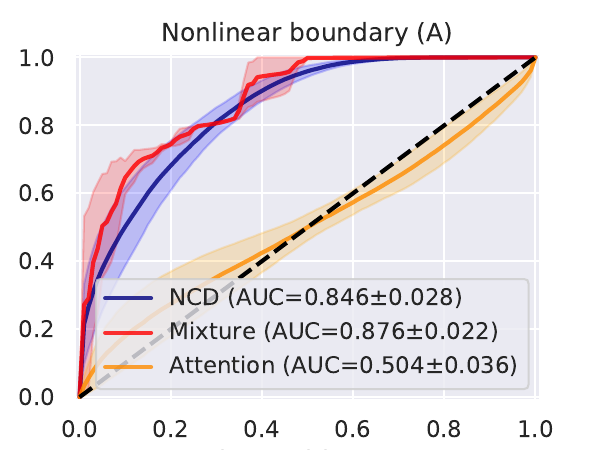}%
\hfill\includegraphics[height=0.25\textwidth]{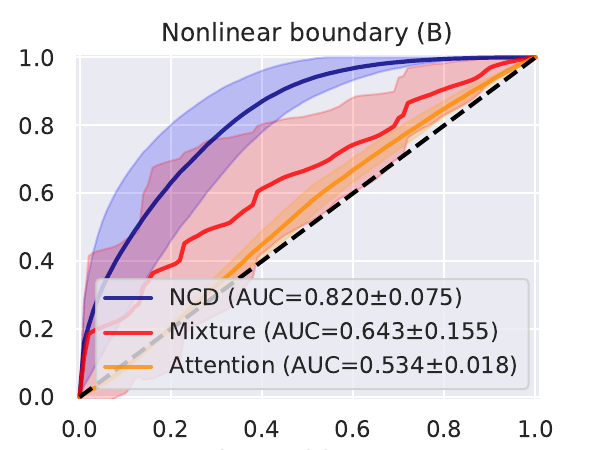}\smallskip

\includegraphics[height=0.25\textwidth]{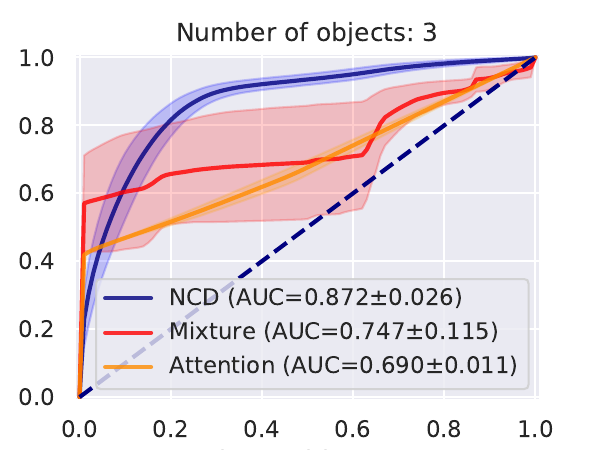}%
\hfill\includegraphics[height=0.25\textwidth]{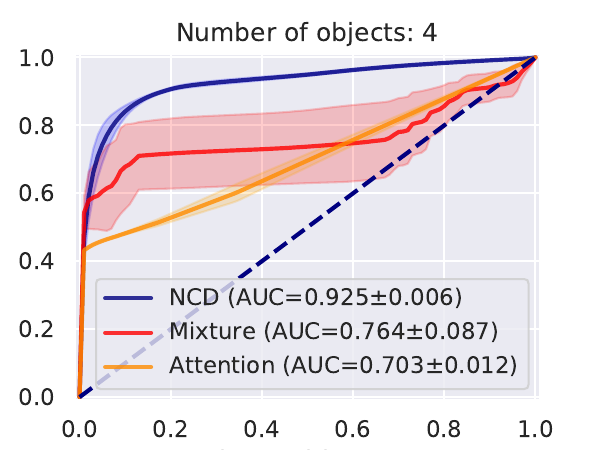}%
\hfill\includegraphics[height=0.25\textwidth]{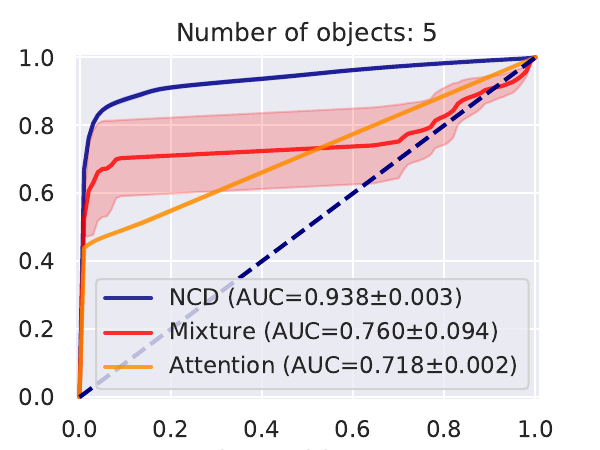}\smallskip

\caption{ROC curves for the ground truth local independence relationships on (top, middle) the synthetic dataset of (top) uniformly and (middle) non-uniformly distributed local parent sets with different types of boundaries and (bottom) on the Spriteworld.}
\label{fig:exp_syn12sprite}
\end{figure*}

We illustrate our results in \Cref{fig:exp_syn12sprite} for (top) the uniformly and (middle) non-uniformly distributed local parent sets, respectively. 
Our method successfully discovers local independence structures and outperforms the baselines in most cases.
We again emphasize that the mixture model exploits the ground truth number of partition sets for its implementation. 

\paragraph{Visualization of the predicted partition.}
We present plots in \Cref{fig:boundary} illustrating how the neural network learns the partition throughout the training procedure based on a toy experiment with 
2 parent variables $\mathbf{X} = \{X_1, X_2\}$.
The quality of the discovered contextual decomposition is improved as the training goes on. 
Our model successfully discovered the local independence and the corresponding context sets. 
It shows that the data points within each partition are properly classified, while the model fails to classify some points close to the decision boundary.

\subsection{Complex System Reflecting Real-world Dynamics}
\label{sec:experiment_spriteworld}
We examine our method on the modified Spriteworld environment \citep{watters2019cobra, pitis2020counterfactual}, which reflects real-world physics dynamics, to evaluate our method in a more complex environment. In a 2 dimensional space, there are $n$ moving objects (i.e., sprites), which can collide with each other. 
Each $i$-th object corresponds to two 2-dimensional variables $X_{2i-1} = (x, y)_{(i)}, X_{2i} = (v_x, v_y)_{(i)}$ which represents the position and the velocity of the object.
An agent can influence the objects through an action.
Action corresponds to a 2-dimensional variable $X_{2n+1} = (a_x, a_y)$, representing some position in the 2D scene, and it influences the dynamics of the nearest object, thus there are interactions between the action and each object. 
In each time-step, the object and action variables influence the object variables of the next time-step.
Formally, $\rmX = \{X_1, \cdots, X_{2n+1}\}$ and $\rmY = \{Y_1, \cdots, Y_{2n}\}$ where $\rmX$ denotes the object and action variables and $\rmY$ denotes the object variables of the next time-step.
Further, there exist dependencies among the parent variables.
Each $Y_k$ is generated from $\rmX$ with an unknown dynamics function $f_k$, i.e., $Y_k = f_k(\rmX, U)$ where $U$ is the exogenous variable.
As a whole, all $\rmX$ are involved in generating each $Y_k$ since the objects can collide with each other and the action influences the objects. 
However, the interactions are sparse---the collision occurs only when objects are (i) close enough and (ii) heading toward each other.
Thus, the dynamics system exhibits numerous \nc{} relationships.
As shown in the bottom of \Cref{fig:exp_syn12sprite}, our method recovers the ground truth local structure on the varying number of objects. 
We provide further details of the dataset and the implementation in \Cref{sec:appendix_spriteworld}.

\section{Related Work}
\label{sec:related_work}

\paragraph{Local independence relationship.}
One of the most widely used local independence relationships is the notion of context-specific independence (CSI) \citep{boutilier2013contextspecific, poole2013contextspecific}. \cite{Pensar_2014} proposed LDAG, a graph labeled with CSIs. In order to represent a more flexible local relationship, \cite{pensar2016role} introduced partial conditional independence (PCI), which is a generalization of CSI. The benefits of exploiting local independence relationships such as CSI has explored recently in the field of causal effect identification. \cite{tikka2020identifying} established the framework of causal calculus in the presence of CSI relationships along with LDAG and showed that it enables richer causal effect identification. \cite{robins2020interventionist} proposed the algorithm for the identification of controlled direct effect leveraging CSI relationships. 
While our focus is on continuous variables in SCM, \cite{nyman2017stratified} also considered continuous variables following Gaussian distributions but discretized the variables with certain intervals. Hence it is limited in terms of both the type of distribution and the conditioning set for local independence.

\paragraph{Local independence of continuous variables.}
\cite{pitis2020counterfactual} proposed attention-based methods to learn local causal structures. They trained a single-head transformer network and an adjacency matrix is then obtained by the product of softmax attention masks of each layer. Since it is trained \textit{without} an inferred local causal graph, its prediction uses all of the values of the input and does not take any local independence into account for approximating conditional distribution. In contrast, our method learns \textit{both} local causal graph within the parental relationship and conditional distribution simultaneously.
Recently, \cite{seitzer2021causal} proposed to estimate the conditional mutual information to discover the local independence relationships. However, they assumed that the underlying ground truth conditional distribution is the Gaussian in order to compute the conditional mutual information. In addition, their method is restricted to a single edge, i.e., one of the parent variables and the target variable, thus is not scalable and does not discover a \nc{} relationships.

\paragraph{Neural network based causal discovery.}
Causal discovery attempts to reconstruct a ground truth causal graph, e.g., through conditional independence induced from the underlying system, where most of the methods fall into one of the three categories: constraint-based \citep{spirtes2000constructing}, score-based \citep{chickering2002optimal,zheng2018dags,yu2019dag,lachapelle2019gradient}, or hybrid \citep{wang2017permutation} (see \citep{glymour2019review} for an extensive review). In contrast, our goal is to discover local independence especially within a variable and its parents.
While masking the input instances with the learned mask (i.e., adjacency matrix) is a widely used technique in prior works on neural network based causal discovery \citep{kalainathan2018sam, ng2019masked, brouillard2020differentiable}, our work has a different focus. The key difference is that they learn the mask which is invariant to individual instances, where there exists a single ground truth adjacency matrix. In our work, a mask is a function of the input variables which is modeled as \nv{} where there exist multiple valid contextual decompositions. For the same reason, they do not require labeling instances, which is essential for valid approximation in our model.

\section{Conclusion}
Local independence relationships (e.g., CSI) could be leveraged in many tasks, but most of the methods are restricted to discrete variables. 
To this end, we formalized the notion of \newcsi{} (\nc) providing a more flexible representation of local independence and further allowing both discrete and continuous variables.
Along with \nc{}, we introduced \textit{local parent set} which allows \nc{} to be represented in a causal graph and hence provides an intuitive and compact description of local independence.
For the discovery of \nc{}s in a system, we focused on continuous variables which is challenging, since it is impractical to directly test every individual \nc{}s due to the nature of continuous variables and adopting prior methods to discover CSIs for discrete variables is not trivial. In this work, we tackled this challenge by finding the \textit{contextual decomposition}, a partition of the joint outcome space where each partition is \nc{}-inducing context-set. 
We presented \ours{}, effectively discovering such decomposition by augmenting an auxiliary \newvariable{} (\nv) which enables the gradient-based training. While finding \nv{} is equivalent to finding contextual decomposition which implies the discovery of \nc{}s, experimental results showed that \ours{} successfully finds \nv{} and discovers \nc{}s in a synthetic dataset and is also effective in a complex environment reflecting real-world dynamics.

\acks{We would like to thank Hyungseok Song and the anonymous reviewers for their valuable feedback. 
This work was supported by the Institute of Information \& Communications Technology Planning \& Evaluation (2021-0-02068-AIHub/10\%, 2021-0-01343-GSAI/10\%, 2022-0-00951-LBA/30\%, 2022-0-00953-PICA/50\%) grant funded by the Korean government.}

\bibliography{mainbib}
\newpage
\appendix

\section{Appendix for Preliminaries}
\label{sec:appendix}

\subsection{Local Independence}
\label{sec:appendix_ci}

CSI is the most widely studied local independence relationship which generalizes the notion of conditional independence (CI). If the causal relationship between variable and its parents exhibits a CSI relationship, it implies that the variable is independent of the subset of its parents given a certain context. Labeled DAG (LDAG) \citep{Pensar_2014} is developed to encode CSI relationships on the underlying DAG, where each edge is labeled with a set of contexts, which invoke CSI relationships. Intuitively speaking, each label on its corresponding edge $(j)$ represents the condition of the edge to be inactive, i.e., $X_j$ does not affect $Y$ given the context in the label.
In the field of causal inference, it has been shown that the presence and knowledge of CSI relationships provide richer causal effect identification \citep{tikka2020identifying}.  
PCI \citep{pensar2016role} extends CSI in that, given the specific context, the local independence relationship only holds in a certain domain (i.e., $\mathcal{D}_{A}$). We would like to note again that the line of works of aforementioned local independences mostly focused on finite discrete variables.

Revisiting \Cref{ex:CSSI1}, CSI relationship $Y \Perp X_2 \mid X_1=c$ holds for every $ c<\sfrac{1}{2}$,
however $Y \not\Perp X_1 \mid X_2=c$ for any $c\in [0,1]$. PCI relationships in this system are as follows:
\begin{align}
\label{eq:pci_1}
&Y \Perp X_2 \mid X_2 \in [0, 1], X_1=c \quad (\forall c<\sfrac{1}{2}), \\
&Y \Perp X_2 \mid X_2 \in [0, \sfrac{1}{2c}], X_1=c \quad (\forall c\geq\sfrac{1}{2}), \\
&Y \Perp X_1 \mid X_1 \in [\sfrac{1}{2c}, 1], X_2=c \quad (\forall c\geq\sfrac{1}{2}).
\end{align}
Note that PCI in \Cref{eq:pci_1} is equivalent to CSI $Y \Perp X_2 \mid X_1=c$.
We now show that \nc{} subsumes CSI and PCI.
\nc{} generalizes CSI and PCI and is compatible with both discrete and continuous variables. 

\begin{restatable}[\nc{} subsumes CSI and PCI]{proposition}{subsume}
\label[proposition]{prop:subsume}
(1) For any CSI relationship $Y \Perp \*{X}_{B} \mid \*{X}_{A} = \*{x}_{A}$, there exists a context set $\gE\subseteq\gX$ such that \nc{} relationship $Y \Perp \*{X}_{B} \mid \*{X}_{A}, \gE$ holds.\\
(2) For any PCI relationship $Y \Perp \*{X}_{B} \mid \mathcal{D}_{B}, \*{X}_{A} = \*{x}_{A}$, there exists a context set $\gE\subseteq \gX$ such that \nc{} relationship $Y \Perp \*{X}_{B} \mid \*{X}_{A}, \gE$ holds.
\end{restatable}
\begin{proof}
(1) By the definition of CSI, $p\left(y \mid \*{x}_{A}, \*{x}_{B}\right)=p\left(y \mid \*{x}_{B}\right)$ holds for all $\*{x}_{A} \in \gX_{A}$ whenever $p\left(\*{x}_{A}, \*{x}_{B}\right)>0$. Let $\gE=\{(\*x_{A}, \*x_{B})\mid \*x_{A} \in \gX_{A}\}\subseteq\gX$. Then, $p\left(y \mid \*{x}_{A}, \*{x}_{B}\right) = p\left(y \mid \*{x}_{B}\right) =p\left(y \mid \*{x}'_{A}, \*{x}_{B}\right)$ for every $\left(\*{x}_{A}, \*{x}_{B}\right), \left(\*{x}'_{A}, \*{x}_{B}\right) \in \gE$. Therefore, $\gE$ is a context set which induces \nc{} relationship $Y \Perp \*{X}_{A} \mid \*{X}_{B}, \gE$. \\
(2) By the definition of PCI, 
$p\left(y \mid \*{x}_{A}, \*{x}_{B}\right)=p\left(y \mid \*{x}'_{A}, \*{x}_{B}\right)$
holds for all $\*{x}_{A}, \*{x}_{A}' \in \mathcal{D}_{A}$ whenever $p\left(\*{x}_{A}, \*{x}_{B}\right), p\left(\*{x}'_{A}, \*{x}_{B}\right)>0$. Let $\gE=\{(\*x_{A}, \*x_{B})\mid \*x_{A} \in \mathcal{D}_{A}\}\subseteq\gX$. It directly follows that $p\left(y \mid \*{x}_{A}, \*{x}_{B}\right) =p\left(y \mid \*{x}'_{A}, \*{x}_{B}\right)$ for every $\left(\*{x}_{A}, \*{x}_{B}\right), \left(\*{x}'_{A}, \*{x}_{B}\right) \in \gE$, thus $\gE$ is a context set which induces \nc{} relationship $Y \Perp \*{X}_{A} \mid \*{X}_{B}, \gE$.
\end{proof}

\section{More general condition for the uniqueness of regular CSSI}
\label{sec:appendix_coordinate}

We adopt and slightly modify the notion of coordinate-wise connectedness from \cite{peters2015intersection}.
\begin{definition}
\label{def:coordinate}
Let $\gE\subseteq \gX$, $\gE|_{\*x_{A}} = \{\*x_{A^c}\mid (\*x_{A}, \*x_{A^c})\in\gE\}\subseteq \gX_{A^c}$ and $\text{proj}_{A}(\gE) = \{\*x_{A}\mid \gE|_{\*x_{A}} \neq \emptyset\} \subseteq \gX_{A}$.
Let $\{\gE|_{\*x_{A}}^{(k)}\}_{k\in\gI}$ be the path-connected components of $\gE|_{\*x_{A}}$. For $\*X_{S}, \*X_{T} \subseteq \*X_{A^c}$, we say $\gE|_{\*x_{A}}^{(i)}$ and $\gE|_{\*x_{A}}^{(j)}$ are {\bf directly coordinate-wise connected} w.r.t $\*X_{S}$ and $\*X_{T}$ if $\text{proj}_{S}(\gE|_{\*x_{A}}^{(i)}) \cap \text{proj}_{S}(\gE|_{\*x_{A}}^{(j)}) \neq \emptyset$ or $\text{proj}_{T}(\gE|_{\*x_{A}}^{(i)}) \cap \text{proj}_{T}(\gE|_{\*x_{A}}^{(j)}) \neq \emptyset$. 
We say $\gE|_{\*x_{A}}$ is {\bf coordinate-wise connected} w.r.t $\*X_{S}$ and $\*X_{T}$ if for any pair of its path-connected components $\gE|_{\*x_{A}}^{(i)}$ and $\gE|_{\*x_{A}}^{(j)}$, there is a sequence $(i=a_0, a_1, \cdots, a_n=j)$ such that $\gE|_{\*x_{A}}^{(a_{t-1})}$ and $\gE|_{\*x_{A}}^{(a_{t})}$ are directly coordinate-wise connected w.r.t $\*X_{S}$ and $\*X_{T}$ for all $t=1, \cdots, n$.
We say $\gE$ is {\bf absolutely coordinate-wise connected} if for any $\rmX_A \subseteq \rmX$ and $\*x_{A}\in \text{proj}_{A}(\gE)$, $\gE|_{\*x_{A}}$ is coordinate-wise connected w.r.t $\rmX_S$ and $\rmX_T$ for any arbitrary $\rmX_S, \rmX_T \subseteq \rmX_{A^c}$.
\end{definition}

The following proposition generalizes \Cref{prop:convexintersection}.

\begin{restatable}[Generalization of \Cref{prop:convexintersection}]{re-proposition}{intersection}
\label[proposition]{prop:intersection}
Suppose \nc{} relationships $Y\Perp \*X_{A^c} \mid \*X_{A}, \gE$ and $Y\Perp \*X_{B^c} \mid \*X_{B}, \gE$ hold. If $\gE|_{\*x_{A\cap B}}$ is coordinate-wise connected w.r.t. $\*X_{A\setminus B}$ and $\*X_{B\setminus A}$ for all $\*x_{A\cap B} \in \text{proj}_{A\cap B}(\gE)$ , then $Y\Perp \*X_{(A\cap B)^c} \mid \*X_{A\cap B}, \gE$ hold.
\end{restatable}
\begin{proof}
The outline of the proof mostly follows the one in \cite{peters2015intersection}. 
It is enough to show that if CSSIs $Y\Perp\*X_{AB} \mid \*X_{CD}, \gE$ and $Y\Perp \*X_{AC} \mid \*X_{BD}, \gE$ hold and $\gE|_{\*x_{D}}$ is coordinate-wise connected w.r.t. $\*X_{B}$ and $\*X_{C}$ for all $\*x_{D} \in \text{proj}_{D}(\gE)$, then $Y\Perp \*X_{ABC} \mid \*X_{D}, \gE$ holds, i.e., 
\begin{equation}
p(y\mid \*x_{A}, \*x_{B}, \*x_{C}, \*x_{D}) = p(y\mid \*x^*_{A}, \*x^*_{B}, \*x^*_{C}, \*x_{D})	
\end{equation}
holds for any $(\*x_{A}, \*x_{B}, \*x_{C}, \*x _{D}), (\*x^*_{A}, \*x^*_{B}, \*x^*_{C}, \*x_{D}) \in \gE$. 
Let $\{\gE|_{\*x_{D}}^{(k)}\}_{k\in\gI}$ be the path-connected components of $\gE|_{\*x_{D}}$. 
First, we will show that for any $\*x_{ABC}, \*x^*_{ABC} \in \gE|_{\*x_{D}}^{(i)}$, $p(y\mid \*x_{A}, \*x_{B}, \*x_{C}, \*x_{D}) = p(y\mid \*x^*_{A}, \*x^*_{B}, \*x^*_{C}, \*x_{D})$ holds. Since $\gE|_{\*x_{D}}^{(i)}$ is path-connected, there exist compact path $\{\rvx^{(t)}_{ABC}\mid 0\leq t\leq 1\}$ such that $\rvx^{(0)}_{ABC} = \rvx_{ABC}$ and $\rvx^{(1)}_{ABC} = \rvx^*_{ABC}$. 
We take the set of $n$ open balls with a small radius $\epsilon$ which is the open cover of the path. 
Let $\rvx^{(a_i)}_{ABC}$ be the center of the $i$-th open ball $B_i$ and suppose $\rvx_{ABC}\in B_1$ and $\rvx^*_{ABC}\in B_n$. 
For any $\rvx'_{ABC}\in B_{j}$ and $\rvx''_{ABC}\in B_{j+1}$, suppose $\rvx'''_{ABC}\in B_{j} \cap B_{j+1}$. Then,
\begin{align*}
p(y\mid \*x'_{A}, \*x'_{B}, \*x'_{C}, \*x_{D}) 
&= p(y\mid \*x'''_{A}, \*x'''_{B}, \*x'_{C}, \*x_{D}) \\
&= p(y\mid \*x'''_{A}, \*x'''_{B}, \*x'''_{C}, \*x_{D}) \\
&= p(y\mid \*x''_{A}, \*x''_{B}, \*x'''_{C}, \*x_{D}) \\
&= p(y\mid \*x''_{A}, \*x''_{B}, \*x''_{C}, \*x_{D})
\end{align*}
hold. 
Therefore, $p(y\mid \*x_{A}, \*x_{B}, \*x_{C}, \*x_{D}) = p(y\mid \*x^*_{A}, \*x^*_{B}, \*x^*_{C}, \*x_{D})$ also holds. 
Now, we will show that for any $(\*x_{A}, \*x_{B}, \*x_{C}) \in \gE|_{\*x_{D}}^{(i)}$ and $(\*x^*_{A}, \*x^*_{B}, \*x^*_{C}) \in \gE|_{\*x_{D}}^{(j)}$, $p(y\mid \*x_{A}, \*x_{B}, \*x_{C}, \*x_{D}) = p(y\mid \*x^*_{A}, \*x^*_{B}, \*x^*_{C}, \*x_{D})$ holds. 
Let $(i=a_0, a_1, \cdots, a_n=j)$ be a sequence such that $\gE|_{\*x_{D}}^{(a_{t-1})}$ and $\gE|_{\*x_{D}}^{(a_{t})}$ are directly coordinate-wise connected w.r.t $\*X_{B}$ and $\*X_{C}$ for all $t=1, \cdots, n$.
For any $\rvx'_{ABC}\in \gE|_{\*x_{D}}^{(a_{j-1})}$ and $\rvx''_{ABC}\in \gE|_{\*x_{D}}^{(a_{j})}$,
$\text{proj}_{B}(\gE|_{\*x_{D}}^{(a_{j-1})}) \cap \text{proj}_{B}(\gE|_{\*x_{D}}^{(a_{j})}) \neq \emptyset$ or $\text{proj}_{C}(\gE|_{\*x_{D}}^{(a_{j-1})}) \cap \text{proj}_{C}(\gE|_{\*x_{D}}^{(a_{j})}) \neq \emptyset$. 
Without the loss of generality, suppose $\exists \*x'''_{B} \in \text{proj}_{B}(\gE|_{\*x_{D}}^{(a_{j-1})}) \cap \text{proj}_{B}(\gE|_{\*x_{D}}^{(a_{j})})$. 
Let $(\*x^p_{A}, \*x'''_{B}, \*x^p_{C}) \in \gE|_{\*x_{D}}^{(a_{j-1})}$ and $(\*x^q_{A}, \*x'''_{B}, \*x^q_{C}) \in \gE|_{\*x_{D}}^{(a_{j})}$.
Then, 
\begin{align*}
p(y\mid \*x'_{A}, \*x'_{B}, \*x'_{C}, \*x_{D}) 
&= p(y\mid \*x^p_{A}, \*x'''_{B}, \*x^p_{C}, \*x_{D}) \\
&= p(y\mid \*x^q_{A}, \*x'''_{B}, \*x^q_{C}, \*x_{D}) \\
&= p(y\mid \*x''_{A}, \*x''_{B}, \*x''_{C}, \*x_{D})
\end{align*}
hold. 
Therefore, $p(y\mid \*x_{A}, \*x_{B}, \*x_{C}, \*x_{D}) = p(y\mid \*x^*_{A}, \*x^*_{B}, \*x^*_{C}, \*x_{D})$ also holds for any $(\*x_{A}, \*x_{B}, \*x_{C}) \in \gE|_{\*x_{D}}^{(i)}$ and $(\*x^*_{A}, \*x^*_{B}, \*x^*_{C}) \in \gE|_{\*x_{D}}^{(j)}$, i.e., it holds for any $(\*x_{A}, \*x_{B}, \*x_{C}, \*x _{D}), (\*x^*_{A}, \*x^*_{B}, \*x^*_{C}, \*x_{D}) \in \gE$. 
Thus, $Y\Perp \*X_{ABC} \mid \*X_{D}, \gE$ holds.
\end{proof}

A sufficient condition for the intersection property of conditional independence (CI) from \cite{peters2015intersection} replaces the strict positiveness. 
In contrast, the intersection property of \nc{} requires strictly positive densities and additional conditions.
The following proposition generalizes \Cref{thm:convexuniqueregular}.

\begin{restatable}[Generalization of \Cref{thm:convexuniqueregular}]{proposition}{uniqueregular}
\label[proposition]{prop:uniqueregular}
Suppose $\gE\subseteq\gX$ is absolutely coordinate-wise connected. For any \nc{} relationship $Y\Perp \*X_{A^c} \mid \*X_{A}, \gE$, there exists an unique $B\subseteq A$ such that regular \nc{} relationship $Y\Perp \*X_{B^c} \mid \*X_{B}, \gE$ holds.
\end{restatable}
\begin{proof}
Suppose regular CSSIs $Y\Perp \*X_{S^c} \mid \*X_{S}, \gE$ and $Y\Perp \*X_{T^c} \mid \*X_{T}, \gE$ hold for some $S$ and $T$. 
Since $\gE$ is absolutely coordinate-wise connected,
$Y\Perp \*X_{(S\cap T)^c} \mid \*X_{S\cap T}, \gE$ holds by \Cref{prop:intersection}.
Since $Y\Perp \*X_{S^c} \mid \*X_{S}, \gE$ and $Y\Perp \*X_{T^c} \mid \*X_{T}, \gE$ are regular, it follows that $S=T=S\cap T$. 
\end{proof}

Unfortunately, it is hard to characterize absolutely coordinate-wise connected sets. 
On the other hand, convex sets are absolutely coordinate-wise connected since a convex subset of $\mathbb{R}^n$ is simply connected and thus path-connected.

\section{Omitted Proofs}
\label{appendix:omitted_proofs}

\regularcssi*
\begin{proof}
(i) By definition, $p\left(y \mid \*{x}_{A}, \*{x}_{A^c}\right)=p(y \mid \*{x}_{A}, \*{x}'_{A^c})$ holds for every $\left(\*{x}_{A}, \*{x}_{A^c}\right), \left(\*{x}_{A}, \*{x}'_{A^c}\right) \in \gE$. 
Suppose $B\supseteq A$.
For every $\*x=\left(\*{x}_{B}, \*{x}_{B^c}\right), \*x'=\left(\*{x}_{B}, \*{x}'_{B^c}\right) \in \gE$, $\*{x}_{A} = \*{x}'_{A}$ holds since $\*{x}_{B} = \*{x}'_{B}$.
Therefore, $p\left(y \mid \*{x}_{B}, \*{x}_{B^c}\right) = p(y\mid \*x) = p\left(y \mid \*{x}_{A}, \*{x}_{A^c}\right) = p\left(y \mid \*{x}_{A}, \*{x}'_{A^c}\right) = p(y\mid \*x') = p(y \mid \*{x}_{B}, \*{x}'_{B^c})$. \\
(ii) Suppose $\gF \subseteq \gE$.
Since $p\left(y \mid \*{x}_{A}, \*{x}_{A^c}\right)=p(y \mid \*{x}_{A}, \*{x}'_{A^c})$ holds for every $\left(\*{x}_{A}, \*{x}_{A^c}\right), \left(\*{x}_{A}, \*{x}'_{A^c}\right) \in \gE$, it also holds for every $\left(\*{x}_{A}, \*{x}_{A^c}\right), \left(\*{x}_{A}, \*{x}'_{A^c}\right) \in \gF$.
\end{proof}

\convexintersection*
\begin{proof}
It directly follows from \Cref{prop:intersection}.
\end{proof}

\convexuniqueregular*

\begin{proof}
It directly follows from \Cref{prop:uniqueregular}.
\end{proof}

\cdatomic*
\begin{proof}
Let $\{(\gE_i, A_i)\}_{i=0}^N$ be a \cd{} such that each \nc{} relationship $Y\Perp \*X_{A_i^c} \mid \*X_{A_i}, \gE_i$ is canonical, $A_i$ is non-identical to each other for all $i$, and $\*X_{A_0} = \*X$.
Let $\{(\gF_j, B_j)\}_{j=0}^M$ be another \cd{} such that each \nc{} relationship $Y\Perp \*X_{B_j^c} \mid \*X_{B_j}, \gF_j$ is canonical and $B_j$ is non-identical to each other for all $j$, and $\*X_{B_0} = \*X$.

(i) Suppose there exist $\gE_i$ and $\gF_j$ such that $p(\gE_i\cap \gF_j)>0$ and $A_i\neq B_j$ for some $i$ and $j$. 
First, we consider the case if $i, j\geq 1$.
Without the loss of generality, suppose that $A_i\setminus B_j \neq \emptyset$.
Since $\gE_i$ and $\gF_j$ are open, we can take a non-empty open ball $\gD \subset \gE_i\cap \gF_j$.
Since $Y\Perp \*X_{A_i^c} \mid \*X_{A_i}, \gD$ and $Y\Perp \*X_{B_j^c} \mid \*X_{B_j}, \gD$ hold, $Y\Perp \*X_{(A_i\cap B_j)^c} \mid \*X_{A_i\cap B_j}, \gD$ holds by \Cref{prop:intersection}.
This contradicts that $Y\Perp \*X_{A_i^c} \mid \*X_{A_i}, \gE_i$ is canonical since $A_i\cap B_j \subsetneq A_i$ and $\gD \subset \gE_i$.
Therefore, $A_i = B_j$.
Now, we consider the case if $i=0$ or $j=0$. 
Without the loss of generality, assume that $i=0$.
Suppose $j\geq 1$.
Since $Y\Perp \*X_{B_j^c} \mid \*X_{B_j}, \gF_j$ holds, $Y\Perp \*X_{B_j^c} \mid \*X_{B_j}, (\gE_0 \cap \gF_j)$ also holds. 
It contradicts that $(\gE_0, A_0)$ is canonical since $(\gE_0 \cap \gF_j) \subset \gE_0$ and $B_j \subsetneq Pa(Y)=A_0$.
Therefore, $j=0$ and $A_0=B_0=Pa(Y)$ trivially holds in this case.

(ii) Now, let $C\subset Pa(Y)$.
Suppose that $p(\gE[C]\bigtriangleup\gF[C])>0$. 
Without the loss of generality, suppose that $p(\gF[C]\setminus\gE[C])>0$.
There exists $\gE_k$ such that $p(\gE_k\cap (\gF[C]\setminus\gE[C]))>0$.
Note that $\gE_k \not\subset \gE[C]$, i.e., $A_k \neq C$.
Therefore, there exists $\gF_t$ such that $\gF_t \subset \gF[C]$ (i.e., $B_t = C$) and $p(\gE_k\cap \gF_t)>0$. 
Therefore, $A_k=B_t=C$. 
However which contradicts that $A_k \neq C$.
\end{proof}

\cdatomicnonidentical*
\begin{proof}
Let $\{(\gE_i, A_i)\}_{i=0}^N$ be a \cd{} such that each \nc{} relationship $Y\Perp \*X_{A_i^c} \mid \*X_{A_i}, \gE_i$ is canonical, $A_i$ is non-identical to each other for all $i$, and $\*X_{A_0} = \*X$.
 Let $\{(\gF_j, B_j)\}_{j=0}^M$ be another \cd{} such that each \nc{} relationship $Y\Perp \*X_{B_j^c} \mid \*X_{B_j}, \gF_j$ is canonical and $B_j$ is non-identical to each other for all $j$, and $\*X_{B_0} = \*X$.
 Suppose there exist $\gE_i$ and $\gF_j$ such that $p(\gE_i\cap \gF_j)>0$ and $A_i\neq B_j$ for some $i, j\geq 1$.
 Without the loss of generality, suppose that $A_i\setminus B_j \neq \emptyset$.
 Since $\gE_i$ and $\gF_j$ are open, we can take a non-empty open ball $\gD \subset \gE_i\cap \gF_j$.
 Since $Y\Perp \*X_{A_i^c} \mid \*X_{A_i}, \gD$ and $Y\Perp \*X_{B_j^c} \mid \*X_{B_j}, \gD$ hold, $Y\Perp \*X_{(A_i\cap B_j)^c} \mid \*X_{A_i\cap B_j}, \gD$ holds by \Cref{prop:intersection}.
 This contradicts that $Y\Perp \*X_{A_i^c} \mid \*X_{A_i}, \gE_i$ is canonical since $A_i\cap B_j \subsetneq A_i$ and $\gD \subset \gE_i$.
 Therefore, $A_i = B_j$.
 Now, suppose that $\gE_i\neq \gF_j$ and $p(\gF_j\bigtriangleup\gE_i)>0$. 
 Without the loss of generality, suppose that $p(\gE_i\setminus\gF_j)>0$.
 There exists $\gE_k$ such that $p(\gE_k\cap (\gF_j\setminus\gE_i))>0$. 
 Since $Y\Perp \*X_{B_j^c} \mid \*X_{B_j}, \gE_k\cap (\gF_j\setminus\gE_i)$ holds, $k\neq 0$.
 Since $(\gE_k\cap (\gF_j\setminus\gE_i)) \subset (\gE_k\cap \gF_j)$, $p(\gE_k\cap \gF_j)>0$ holds and thereby $A_k=B_j$ holds. 
 Thus, $A_i=B_j=A_k$ and it contradicts that $A_i$ is non-identical to each other for all $i$. 
 Therefore, if $p(\gE_i\cap \gF_j)>0$, then $p(\gE_i\bigtriangleup \gF_j)=0$.
\end{proof}

\begin{restatable}{proposition}{cssiandcsi}
\label[proposition]{prop:cssiandcsi}
A \nc{} relationship $Y\Perp \*X_{A^c} \mid \*X_{A}, \gE$ holds if and only if CSI relationship $Y\Perp \*X_{A^c} \mid \*X_{A}, Z=0$ holds, where $Z$ is the deterministic binary random variable defined as $Z=0$ if $\*X\in\gE$ and $Z=1$ otherwise. 
\end{restatable}
\begin{proof}
$(\xrightarrow{})$.
Suppose $Y \Perp \*{X}_{A^c} \mid \*{X}_{A}, \gE$ holds. By definition, $p\left(y \mid \*{x}_{A}, \*{x}_{A^c}\right)=p(y \mid \*{x}_{A}, \*{x}'_{A^c})$ holds for every $\left(\*{x}_{A}, \*{x}_{A^c}\right), \left(\*{x}_{A}, \*{x}'_{A^c}\right) \in \gE$. 
Since $p(\*{x}_{A}, \*{x}_{A^c}, z_0) > 0$ implies $\*{x}\in \gE$ and $\*{x}_{A^c} \in \gE|_{\rvx_{A}}$, we have 
\begin{align*}
p(y\mid \*{x}_{A}, z_0)
&= \int_{\gE|_{\rvx_{A}}} p(y, \*{x}'_{A^c} \mid \*{x}_{A}, z_0) d\*{x}'_{A^c} \\
&= \int_{\gE|_{\rvx_{A}}} p(y\mid \*{x}_{A}, \*{x}'_{A^c}, z_0) p(\*{x}'_{A^c} \mid \*{x}_{A}, z_0) d\*{x}'_{A^c} \\
&= \int_{\gE|_{\rvx_{A}}} p(y\mid \*{x}_{A}, \*{x}'_{A^c}) p(\*{x}'_{A^c} \mid \*{x}_{A}, z_0) d\*{x}'_{A^c} \\
&= \int_{\gE|_{\rvx_{A}}} p(y\mid \*{x}_{A}, \*{x}_{A^c}) p(\*{x}'_{A^c} \mid \*{x}_{A}, z_0) d\*{x}'_{A^c} \\
&= p(y\mid \*{x}_{A}, \*{x}_{A^c}) \int_{\gE|_{\rvx_{A}}} p(\*{x}'_{A^c} \mid \*{x}_{A}, z_0) d\*{x}'_{A^c} \\
&= p(y\mid \*{x}_{A}, \*{x}_{A^c}) = p(y\mid \*{x}_{A}, \*{x}_{A^c}, z_0),
\end{align*}
for every $\left(\*{x}_{A^c}, \*{x}_{A}\right) \in \gE$. Therefore, $Y \Perp \*{X}_{A^c} \mid \*{X}_{A}, Z=z_0$ holds.\\
$(\xleftarrow{})$. 
Suppose $Y \Perp \*{X}_{A^c} \mid \*{X}_{A}, Z=z_0$ holds. By definition, $p(y\mid \*{x}_{A}, \*{x}_{A^c}, z_0) = p(y\mid \*{x}_{A}, z_0)$ holds for every $\*{x}$ with $p(\*{x}_{A}, \*{x}_{A^c}, z_0) > 0$. Since $p(\*{x}_{A}, \*{x}_{A^c}, z_0)=p(\*{x}, z_0) > 0$ is equivalent to $\*{x}\in\gE$, for every $(\*{x}_{A}, \*{x}_{A^c}), (\*{x}_{A}, \*{x}'_{A^c}) \in \gE$ we have 
\begin{align*}
p\left(y \mid \*{x}_{A}, \*{x}_{A^c}\right)
&= p(y \mid \*{x}_{A}, \*{x}_{A^c}, z_0) \\
&= p(y \mid \*{x}_{A}, z_0) \\
&= p(y \mid \*{x}_{A}, \*{x}'_{A^c}, z_0) \\
&= p\left(y \mid \*{x}_{A}, \*{x}'_{A^c}\right).
\end{align*}
Therefore, $Y \Perp \*{X}_{A^c} \mid \*{X}_{A}, \gE$ holds.
\end{proof}

\cdandcsi*
\begin{proof}
 It is the direct extension of \Cref{prop:cssiandcsi}.
\end{proof}

\section{Additional Examples}
\label{sec:appendix_examples}

\begin{example}[Non-existence of non-trivial CD]
\label{ex:existence}
Let $X_1, X_2 \sim \text{Unif }[0, 1]$ and $U$ be an exogenous variable. Let $Y = X_1 + X_2 + U$. In this case, there exists a unique contextual decomposition which is trivial: $\gE = \gX$, i.e., the causal mechanism does not exhibit any \nc{} relationship.
\end{example}

\begin{example}[Augmented Causal Graph, Continued from \Cref{ex:CSSI1}]
\label{ex:CIPV3}
Let $X_1, X_2,$ and $Y$ be defined as \Cref{ex:CSSI1}. Let $Z$ be the binary random variable which is $0$ if $X_1X_2<\sfrac{1}{2}$ and otherwise $1$. Then, the following holds:
\[
Y \Perp X_2 \mid X_1, Z=0, \quad
Y \Perp X_1 \mid X_2, Z=1.
\]
Here, $\{\gE_1, \gE_2\}$ is a contextual decomposition where $\gE_k$ is a support set of $P(Z=k\mid X)$.
\end{example}
We illustrated the implementation of \nv{} in \Cref{fig:example-d,fig:example-e,fig:example-f}. \Cref{fig:example-d} shows an augmented SCM in \Cref{ex:CIPV3}. \Cref{fig:example-e} represents $Y \Perp X_2 \mid X_1, Z=0$, and \Cref{fig:example-f} represents $Y \Perp X_1 \mid X_2, Z=1$.

\begin{figure*}[t!]
\centering\includegraphics[height=0.25\textwidth]{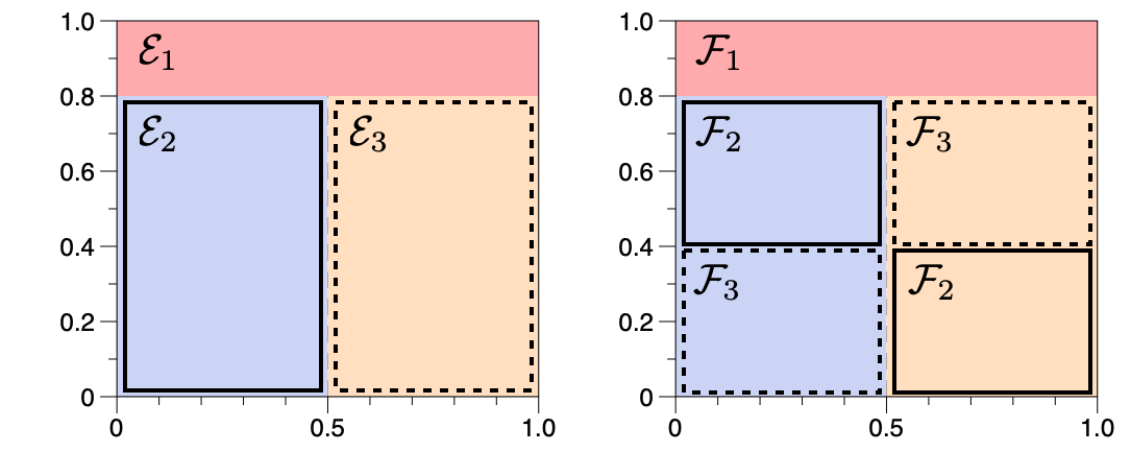}
\caption{\Cref{ex:canonical_cd}.}
\label{fig:example_canonical_cd}
\end{figure*}
\index{figure}

\begin{example}[Canonical \cd{}, Illustrating \Cref{theorem:cdatomic}]
\label{ex:canonical_cd}
Let $X_1, X_2 \sim \textrm{Unif }[0, 1]$ and $U$ be an exogenous variable. 
Let $Y$ be: 
\begin{align*}
Y = 
\begin{cases}
X_1 + U & \text {if }\quad \,\,\,X_2 \geq 0.8, \\
X_2 + U & \text {if }\quad \,\,\,X_1 < 0.5 \,\,\text{and } X_2 < 0.8, \\
X_2 + 2U & \text {if }\quad \,\,\,X_1 \geq 0.5 \,\,\text{and } X_2 < 0.8.
\end{cases}
\end{align*}
Let $\gE_1=\{(x_1, x_2) \mid x_2 >0.8 \}$, $\gE_2=\{(x_1, x_2) \mid x_1<0.5, x_2 <0.8 \}$, $\gE_3=\{(x_1, x_2) \mid x_1>0.5, x_2 <0.8 \}$, and $\gE_0 = (\gE_1 \cup \gE_2 \cup \gE_3)^c$.
Then, canonical \nc{}s $(Y \Perp X_2 \mid X_1, \gE_1)$, $(Y \Perp X_1 \mid X_2, \gE_2)$, and $(Y \Perp X_1 \mid X_2, \gE_3)$ hold. 
Therefore, $\{(\gE_0, \rmX), (\gE_1, \rmX_1), (\gE_2, \rmX_2), (\gE_3, \rmX_2)\}$ is a canonical \cd{}.
On the other hand, let $\gF_1=\gE_1$, $\gF_2=\{(x_1, x_2) \mid (x_1<0.5, 0.4<x_2 <0.8) \text{ or } (x_1>0.5, x_2 <0.4)\}$, $\gF_3=\{(x_1, x_2) \mid (x_1<0.5, x_2 <0.4) \text{ or } (x_1>0.5, 0.4<x_2 <0.8)\}$, and $\gF_0 = (\gF_1 \cup \gF_2 \cup \gF_3)^c$.
Then, canonical \nc{}s $(Y \Perp X_2 \mid X_1, \gF_1)$, $(Y \Perp X_1 \mid X_2, \gF_2)$, and $(Y \Perp X_1 \mid X_2, \gF_3)$ hold. 
Therefore, $\{(\gF_0, \rmX), (\gF_1, \rmX_1), (\gF_2, \rmX_2), (\gF_3, \rmX_2)\}$ is a canonical \cd{}.
Here, $p(\gE_2\cap \gF_2)>0$ and thus $\*{pa}^{\gE_2} = \*{pa}^{\gF_2} = \rmX_2$ by \Cref{theorem:cdatomic}. Also, $\gE[\rmX_2] = \gE_2\cup \gE_3 = \gF_2\cup \gF_3 = \gF[\rmX_2]$ by \Cref{theorem:cdatomic}.
\end{example}

\section{Experimental Details}
\label{sec:appendix_experiment}

\subsection{Synthetic datasets}
\label{sec:appendix_synthetic} 

We provide the details of the configurations of the causal system in our experiments.
The following causal system is the case of uniformly distributed local parent sets with a linear boundary. Here, $g$ is a linear function implemented with a randomly initialized NN.
\begin{align}
Y = 
\begin{cases}
f_1(\rmX_{123}, N) & \text {if }\quad \,\,\, g(\*{x}_{147}) = \max (g(\*{x}_{147}), g(\*{x}_{258}), g(\*{x}_{369})) , \\
f_2(\rmX_{456}, N) & \text {if }\quad \,\,\, g(\*{x}_{258}) = \max (g(\*{x}_{147}), g(\*{x}_{258}), g(\*{x}_{369})), \\
f_3(\rmX_{789}, N) & \text {if }\quad \,\,\, g(\*{x}_{369}) = \max (g(\*{x}_{147}), g(\*{x}_{258}), g(\*{x}_{369})).
\end{cases}
\end{align}
The following is the case of uniformly distributed local parent sets with a nonlinear boundary determined by the norm:
\begin{align}
Y = 
\begin{cases}
f_1(\rmX_{123}, N) & \text {if }\quad \,\,\, \|\*{x}\| < c_1 , \\
f_2(\rmX_{456}, N) & \text {if }\quad \,\,\, c_1 < \|\*{x}\| < c_2, \\
f_3(\rmX_{789}, N) & \text {if }\quad \,\,\, \|\*{x}\| > c_2.
\end{cases}
\end{align}
The following is the case of uniformly distributed local parent sets with a nonlinear boundary determined by a linear function $g$ implemented with a NN:
\begin{align}
Y = 
\begin{cases}
f_1(\rmX_{123}, N) & \text {if }\quad \,\,\, g(\*{x}_{147}) = \max (g(\*{x}_{147}), g(\*{x}_{258}), g(\*{x}_{369})) , \\
f_2(\rmX_{456}, N) & \text {if }\quad \,\,\, g(\*{x}_{258}) = \max (g(\*{x}_{147}), g(\*{x}_{258}), g(\*{x}_{369})), \\
f_3(\rmX_{789}, N) & \text {if }\quad \,\,\, g(\*{x}_{369}) = \max (g(\*{x}_{147}), g(\*{x}_{258}), g(\*{x}_{369})).
\end{cases}
\end{align}
The following causal system is the case of non-uniformly distributed local parent sets with a linear boundary. Here, $g$ is a linear function implemented with a randomly initialized NN.
\begin{align}
Y = 
\begin{cases}
f_1(\rmX_{123}, N) & \text {if }\quad \,\,\, g(\*{x}_{147}) = \max (g(\*{x}_{147}), g(\*{x}_{258}), g(\*{x}_{369})) , \\
f_2(\rmX_{456789}, N) & \text {if }\quad \,\,\, g(\*{x}_{258}) = \max (g(\*{x}_{147}), g(\*{x}_{258}), g(\*{x}_{369})), \\
f_3(\rmX, N) & \text {if }\quad \,\,\, g(\*{x}_{369}) = \max (g(\*{x}_{147}), g(\*{x}_{258}), g(\*{x}_{369})).
\end{cases}
\end{align}
The following is the case of non-uniformly distributed local parent sets with a nonlinear boundary determined by the norm:
\begin{align}
Y = 
\begin{cases}
f_1(\rmX_{123}, N) & \text {if }\quad \,\,\, \|\*{x}\| < c_1 , \\
f_2(\rmX_{456789}, N) & \text {if }\quad \,\,\, c_1 < \|\*{x}\| < c_2, \\
f_3(\rmX, N) & \text {if }\quad \,\,\, \|\*{x}\| > c_2.
\end{cases}
\end{align}
The following is the case of non-uniformly distributed local parent sets with a nonlinear boundary determined by a nonlinear function $g$ implemented with a NN:
\begin{align}
Y = 
\begin{cases}
f_1(\rmX_{123}, N) & \text {if }\quad \,\,\, g(\*{x}_{147}) = \max (g(\*{x}_{147}), g(\*{x}_{258}), g(\*{x}_{369})) , \\
f_2(\rmX_{456789}, N) & \text {if }\quad \,\,\, g(\*{x}_{258}) = \max (g(\*{x}_{147}), g(\*{x}_{258}), g(\*{x}_{369})), \\
f_3(\rmX, N) & \text {if }\quad \,\,\, g(\*{x}_{369}) = \max (g(\*{x}_{147}), g(\*{x}_{258}), g(\*{x}_{369})).
\end{cases}
\end{align}

For the toy experiment for the visualization of the decision boundary, we let $X_1, X_2 \sim \mathcal{N}(0,I_3)$ and $Y = f_1(X_1)$ if $\|(X_1, X_2)\| < \epsilon$ and $f_2(X_2)$ otherwise.
For the linear function $g$, we used a randomly initialized linear layer.
For the nonlinear function $g$, we used a randomly initialized NN with a single hidden layer, 10 hidden units, and Tanh activation. 
The total number of data samples is 50000, 
with a ratio of 8:1:1 of training, validation, and held-out test set, respectively.

\begin{figure*}[t!]
\centering
\includegraphics[width=0.20\textwidth]{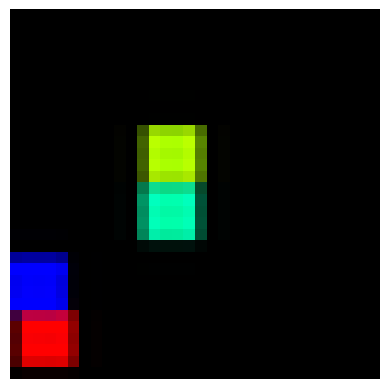}%
\hspace{10pt}
\includegraphics[width=0.20\textwidth]{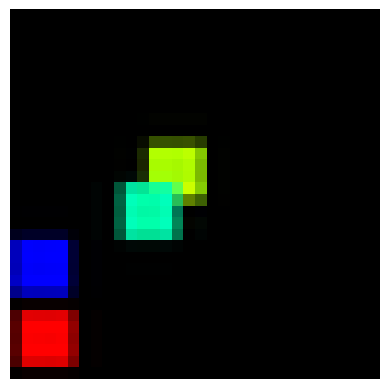}%
\hspace{10pt}
\includegraphics[width=0.20\textwidth]{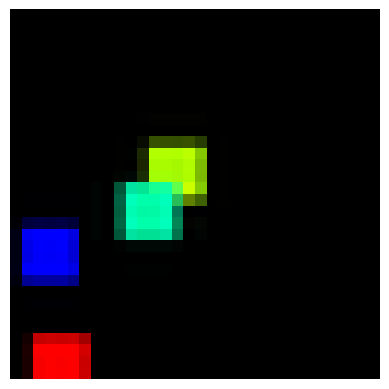}\smallskip
\caption{Example images of the Spriteworld environment.}
\label{fig:dataset_spriteworld}
\end{figure*}

\subsection{Spriteworld}
\label{sec:appendix_spriteworld}
While the original environment does not consider the interaction between objects, \cite{pitis2020counterfactual} modified the environment to devise the collision between moving objects and to acquire the ground truth local causal graphs. 
Our experiment was also conducted in the modified Spriteworld, which we denote as Spriteworld for simplicity. 
In this environment, there are moving objects in the scene. 
Following real-world dynamics, objects can collide with each other, and thus each object has an influence on others. However, their position and velocity will be affected by the others only when they collide, which occurs relatively rarely. In other words, the state of each object will only be determined by its own state in the previous time-step in most cases. 
Example images of the Spriteworld environment are shown in \Cref{fig:dataset_spriteworld}.
In the setting of \cite{pitis2020counterfactual}, each variables represent corresponding objects, i.e., $X_i=(x, y, v_x, v_y)_{i}$ represents the $i$-th object. 
Thus, $\rmX = \{X_1, \cdots, X_{n+1}\}$ and $\rmY = \{Y_1, \cdots, Y_{n}\}$.
In our experiment, we consider a more complex setup where each variable represents the position and velocity of each object, thus $\rmX = \{X_1, \cdots, X_{2n+1}\}$ and $\rmY = \{Y_1, \cdots, Y_{2n}\}$. Further, we also include the exogenous variable.

\subsection{Implementation.}
\label{sec:appendix_implementation}
\paragraph{Implementation details.}
For the implementation of our model, the distribution $p_\theta(y | x, z)$ could be any distribution as long as the log-likelihood is computable, and we chose Gaussian for simplicity. 
We note that ground truth $p(y | x, z)$ is non-Gaussian (i.e., non-linear function with non-additive noise model) in our experiments.
For $f_{\theta}(\mathbf{x}, z)$, which approximates the conditional distribution $P(Y\mid \mathbf{X}, Z)$, and $g_{\phi}(\mathbf{x})$, which outputs the parameters of Bernoulli distribution, we used an MLP with 3 hidden layers and 128 hidden units. 
For all the experiments, we set the batch size to 1000 and used the Adam optimizer with the weight decay of $10^{-5}$. 
We set the learning rate of $10^{-2}$ for the synthetic dataset and $10^{-3}$ for the Spriteworld. 
Most experiments were conducted with a single RTX 3090 GPU.
For the mixture model and the transformer model, we used 3 MLPs, each with 3 hidden layers and 128 hidden units, and grid-search the learning rate over $\{0.005, 0.01, 0.02, 0.03\}$.
\paragraph{Regularizer.}
One can adopt a regularizer to induce a sparse solution and prevent the model from outputting a trivial solution $z=(1, 1, \cdots, 1)$ where $z\sim g_\phi(\mathbf{x})$, e.g., adding a term $\lambda\cdot \|g_\phi(\mathbf{x})\|_1$. 
During our experiments, however, we empirically found that 
the regularizer does not significantly bring a gain.
We hypothesized that since the neural network $\phi$ is randomly initialized, approximately half of the entries of $z$ would be $0$ at the beginning of the training, and, consequently, it would rarely converge to the trivial solution $z=(1, 1, \cdots, 1)$.
\paragraph{Gumbel-Softmax.}
Although there are other methods used for learning discrete variables, such as a REINFORCE estimator, we chose Gumbel-Softmax \citep{jang2016categorical,maddison2016concrete} since it (i) has a lower variance compared to REINFORCE estimator, and (ii) is empirically shown to be more effective \citep{jang2016categorical,maddison2016concrete,ng2019masked}.
Given that the choice of reparametrization trick is not the main focus of our work, we did not compare it to other estimators.
\paragraph{Evaluation.}
We now describe the procedure for plotting the ROC curve.
Let $Pa(Y)=\{1,2, \cdots ,9\}$ (i.e., $d=9$). 
Consider a \textit{single data point} $\mathbf{x}$ where $A_{true}=\{1\}$, and the model predicted $A_{pred}=\{1, 2\}$ with a threshold $\tau$. 
In this case, $TP=1$, $FP=1$, $FN=0$, and $TN=7$. 
We note that this evaluation procedure follows the prior work \citep{pitis2020counterfactual}.
We plot the ROC curves based on \textit{all the data points} in a test dataset. 
Accordingly, $TP+FP+FN+TN = B\times d$ where $B$ is the number of test data points. 
Each data point belongs to one of the partitioned sets (i.e., region), and the number of data points for each region is similar. 
Train and test distributions are identical.

\section{Additional Experiments}
\label{sec:appendix_additional_experiment}

\begin{figure*}[t!]
\centering\includegraphics[height=0.25\textwidth]{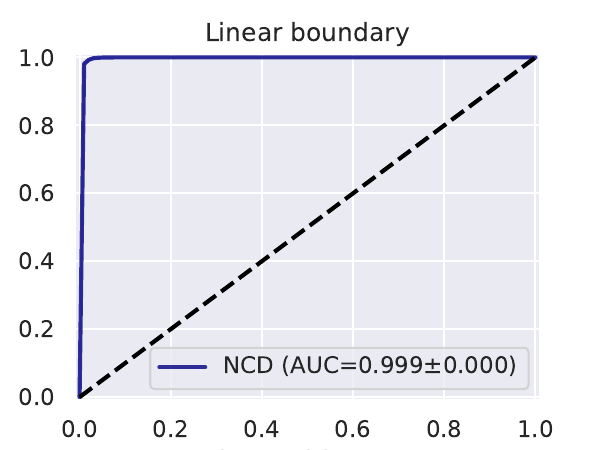}%
\hfill\includegraphics[height=0.25\textwidth]{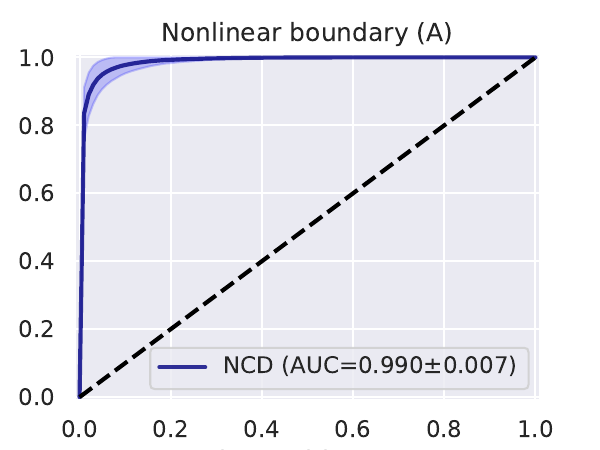}%
\hfill\includegraphics[height=0.25\textwidth]{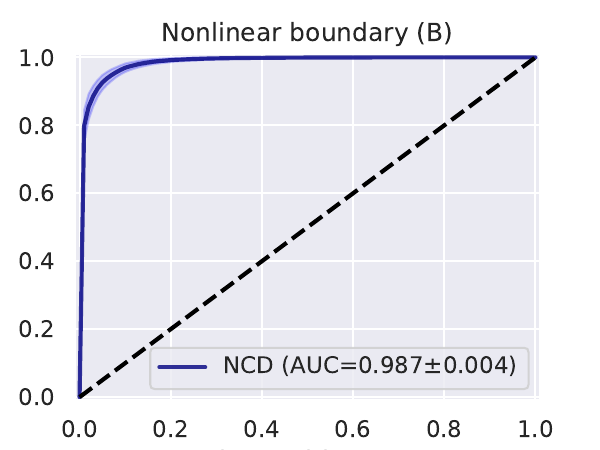}\smallskip

\caption{ROC curves for the ground truth local independence relationships on the synthetic dataset of the additive noise model.}
\label{fig:exp_syn_nnadd}
\end{figure*}

Here, we provide the experimental results on the synthetic datasets where the functions are nonlinear with an additive noise model. 
We consider the following causal system with a linear boundary determined by a linear function $g$:
\begin{align}
Y = 
\begin{cases}
f_1(\rmX_{123}) + N & \text {if }\quad \,\,\, g(\*{x}_{147}) = \max (g(\*{x}_{147}), g(\*{x}_{258}), g(\*{x}_{369})) , \\
f_2(\rmX_{456}) + N & \text {if }\quad \,\,\, g(\*{x}_{258}) = \max (g(\*{x}_{147}), g(\*{x}_{258}), g(\*{x}_{369})), \\
f_3(\rmX_{789}) + N & \text {if }\quad \,\,\, g(\*{x}_{369}) = \max (g(\*{x}_{147}), g(\*{x}_{258}), g(\*{x}_{369})).
\end{cases}
\end{align}
The following is the case of a nonlinear boundary determined by the norm:
\begin{align}
Y = 
\begin{cases}
f_1(\rmX_{123}) + N & \text {if }\quad \,\,\, \|\*{x}\| < c_1 , \\
f_2(\rmX_{456}) + N & \text {if }\quad \,\,\, c_1 < \|\*{x}\| < c_2, \\
f_3(\rmX_{789}) + N & \text {if }\quad \,\,\, \|\*{x}\| > c_2.
\end{cases}
\end{align}
The following is the case of a nonlinear boundary determined by a nonlinear function $g$:
\begin{align}
Y = 
\begin{cases}
f_1(\rmX_{123}) + N & \text {if }\quad \,\,\, g(\*{x}_{147}) = \max (g(\*{x}_{147}), g(\*{x}_{258}), g(\*{x}_{369})) , \\
f_2(\rmX_{456}) + N & \text {if }\quad \,\,\, g(\*{x}_{258}) = \max (g(\*{x}_{147}), g(\*{x}_{258}), g(\*{x}_{369})), \\
f_3(\rmX_{789}) + N & \text {if }\quad \,\,\, g(\*{x}_{369}) = \max (g(\*{x}_{147}), g(\*{x}_{258}), g(\*{x}_{369})).
\end{cases}
\end{align}

As shown in \Cref{fig:exp_syn_nnadd}, our method successfully discovers the local independence relationships in the additive noise system as well.

\end{document}